\documentclass{article}
\PassOptionsToPackage{round}{natbib}
\usepackage[final]{neurips_2025}
\usepackage{style_weikai}

\title{Planning and Learning in Average Risk-aware MDPs}

\author{Weikai Wang\textsuperscript{1,2} \hspace{1.5cm} Erick Delage\textsuperscript{1,2}\\
\textsuperscript{1} GERAD \& HEC Montréal \hspace{1cm} \textsuperscript{2}Mila - Québec AI Institute\\
\texttt{\{weikai.wang, erick.delage\}@hec.ca}
}

\begin{document}

\maketitle

\begin{abstract}
For continuing tasks, average cost Markov decision processes have well-documented value and can be solved using efficient algorithms. However, it explicitly assumes that the agent is risk-neutral. In this work, we extend risk-neutral algorithms to accommodate the more general class of dynamic risk measures. Specifically, we propose a relative value iteration (RVI) algorithm for planning and design two model-free Q-learning algorithms, namely a generic algorithm based on the multi-level Monte Carlo (MLMC) method, and an off-policy algorithm dedicated to utility-based shortfall risk measures. Both the RVI and MLMC-based Q-learning algorithms are proven to converge to optimality. Numerical experiments validate our analysis, confirm empirically the convergence of the off-policy algorithm, and demonstrate that our approach enables the identification of policies that are finely tuned to the intricate risk-awareness of the agent that they serve.
\end{abstract}

\section{Introduction}
For continuing tasks where there is a need to optimize a long term periodic payoff, such as network control, supply chain designs, or maintenance problems \citep{Puterman1994book},  average cost (or reward) Markov decision processes (MDPs) serve as a crucial model in reinforcement learning \citep{Sutton2018book, Naik2019arXiv} and can be solved using efficient algorithms. In the risk-neutral setting, different forms of value iteration algorithms co-exist \citep{Puterman1994book, Bertsekas2007book} and some have been extended to a model-free setting using Q-learning style algorithms \citep{Abounadi2001sicon, Wan2021icml}. The question of how to formulate and solve average-cost MDPs however becomes challenging when the agent is considered risk sensitive. It originates from the pioneering work of \cite{Howard1972} and is covered in recent surveys such as \cite{Biswas2023} and \cite{Bauerle2024mmor}.

This work focuses on average risk-aware MDPs, a general formulation introduced in \cite{Shen2013sicon} that attempts to find a policy $\bm{\pi}$ that minimizes the long-term average of the risk of the cost process generated by $\bm{\pi}$. 
While the theoretical foundations of this framework are well studied, finding efficient solution techniques to these problems remains a challenging task.
To solve the average risk-aware MDP problem, one can apply the classic value iteration algorithm \citep{Cavazos2003mor, Ruszczynski2010, Shen2015arXiv}. This approach relies on the iteration of the risk-aware Bellman operator and computes the average to obtain the optimal average risk. However, it is known to suffer from overflow issues when the number of iterations is large. In the risk-neutral setting, the relative value iteration (RVI) algorithm is widely used \citep{Bertsekas2007book, Gupta2015cdc}, as it mitigates overflow issues during long iterations by subtracting a reference value for each state at every step. However, while some studies have explored RVI algorithms based on the entropic risk measure \citep{Borkar2010mtns, Arapostathis2019, Hmedi2023}, a general formulation of the RVI algorithm for risk-aware MDPs remains missing in the literature. 

Meanwhile, in practical applications, the environment is rarely known in full, highlighting the importance of developing model-free learning algorithms.
To the best of our knowledge, extensions of the risk-neutral Q-learning algorithms to the average risk-aware setting appear to only exist for the case of entropic risk measure (see \cite{Borkar2002mor}, \cite{Borkar2010mtns},  \cite{Moharrami2024mor} and the reference therein). This is in sharp contrast to the extensive literature on algorithms for discounted or finite-horizon risk-aware MDPs, where many studies exist: for instance, see 
\cite{Chow2014neurips}, \cite{Tamar2015aaai}, or \cite{Chow2018jmlr} for conditional value-at-risk, see \cite{Huang2017cdc}, \cite{Kose2021jmlr}, or \cite{Lam2023iclr} for general coherent risk measures, \cite{Shen2014neuralcomp} or \cite{Marzban2023} for utility-based shortfall risk (UBSR), and see  
\cite{Hau2024aistats} for quantiles. 
To conclude, the design of a model-free learning algorithm for average risk-aware MDPs with a general risk measure remains an open research field.

The literature most closely related to addressing this gap has focused on planning and learning algorithms for distributionally robust MDPs, leveraging the fact that coherent risk measures admit a worst-case expected value representation. Studies have explored the discounted case \citep{Liu2022icml, WangSB2023aistats, WangSB2024jmlr} as well as the average case \citep{WangY2023aaai, WangY2023icml}, where an ambiguity set is constructed around the transition kernel to safeguard against potential distributional shifts. These results, however, do not apply to general classes of possibly non-coherent dynamic risk measures. 

This paper presents planning and learning algorithms for average risk-aware MDPs with a general dynamic risk measure. We describe our contributions as follows: 
\begin{enumerate}
    \item \textbf{Planning:} We propose a model-based RVI algorithm for average risk-aware MDPs, which produces a policy that provably converges to the optimal policy for a general class of dynamic risk measures. While existing studies on model-based algorithms for this problem focus either on the risk-neutral setting or the case of entropic risk, our work appears to be the first to consider such a general class of dynamic risk measures. 
    \item \textbf{Learning:} We introduce two novel model-free Q-learning algorithms for average risk-aware MDPs. The first one generalizes the multi-level Monte Carlo (MLMC) based Q-learning algorithm introduced in \cite{WangY2023icml} for robust average MDPs to a broader class of dynamic risk measures, which may not necessarily be coherent, while ensuring provable convergence to optimality. The conditions that we impose for convergence are weak and are satisfied by many popular risk measures such as UBSR, optimized certainty equivalent, and spectral risk measures. Additionally, we propose an asynchronous algorithm that is specialized for UBSR and amenable to off-policy learning by waiving the need for a resampling procedure. While the theoretical convergence remains open, we validate it empirically under different loss functions. 

    \item \textbf{Empirics:} We confirm empirically the convergence of all algorithms under different choice of risk measures and practically relevant sampling rates for MLMC Q-learning, and compare the sample efficiency. We also showcase how average risk-aware MDPs identify policies that are tuned to the agents' preferences in popular environments from the literature.
\end{enumerate}

The structure of the paper is as follows. Section \ref{sec-Preliminaries} introduces average risk-aware MDPs. 
Section \ref{sec-ModelBased} introduces the model-based algorithms, including the risk-aware RVI and its generalization to Q-factors. Section \ref{sec-ModelFree} describes the model-free algorithms, featuring a general Q-learning algorithm with MLMC and an asynchronous Q-learning algorithm specifically designed for UBSR. Section \ref{sec-Experiments} presents the numerical experiments. Section \ref{sec-Conclusion} concludes the paper and proposes further research. Pseudo-codes, proofs, and additional experiment details and results are provided in the appendix.

\section{Preliminaries}\label{sec-Preliminaries}
\textit{Notations:} Given any finite probability space $(\Omega, \sigma(\Omega), P(\cdot))$, abbreviated as $(\Omega, P(\cdot))$,
with $\Omega$ a finite set of outcomes, $\sigma(\Omega)$ the power set of $\Omega$ (sigma-algebra) and $P(\cdot)$ a probability mass function in the probability simplex $\mathscr{P}(\Omega)$, we denote by $\mathcal{L}(\Omega)$ the set of finite real-valued functions (a.k.a. random variables) on $\Omega$ and $|\Omega|$ the cardinality of $\Omega$.
For $v, w \in \mathcal{L}(\Omega)$, the notation $v \geq w$ refers to $v(\omega) \geq w(\omega)$ for all $\omega \in \Omega$, and $v \geq w$ almost surely (a.s.) refers to $v(\omega) \geq w(\omega)$ for all $\omega \in \Omega$ such that $P(\omega)>0$. 
The infinity norm of $v \in \mathcal{L}(\Omega)$ is $\|v\|_\infty := \sup_{\omega \in \Omega} |v(\omega)|$, while its span-seminorm is: $\|v\|_{sp} := \max_{\omega \in \Omega} v(\omega) - \min_{\omega \in \Omega} v(\omega)$.
For $A \subseteq  \Omega$, the indicator function $\bm{1}\{\omega \in A\}$ equals $1$ if $\omega \in A$ and $0$ otherwise. 
Finally, $e$ and $\bm{0}$ represent the constant functions of one and zero respectively, while $\mathrm{e}$ denotes the base of the natural logarithm. 

\subsection{Risk Maps}
We begin by defining the notion of a risk measure following \cite{Shapiro2021book}.
\begin{defn}
    Given a finite probability space $(\Omega,P(\cdot))$, a risk measure $\rho: \mcL(\Omega) \to \mbR$ that maps a random cost to a real value capturing its risk is said to be monetary if it satisfies the following properties:
    \begin{enumerate}[label=(\arabic*)]
        \item (Monotonicity) $\rho(v) \le \rho(w)$ for all $v,w \in \mcL(\Omega)$ such that $v\leq w$ a.s.;
        \item (Translation invariance) $\rho(v+\lambda) = \rho(v) + \lambda$ for any $\lambda \in \mbR$, $v\in\mcL(\Omega)$;
        \item (Normalization) $\rho(0) = 0$;
    \end{enumerate}
    further called convex if:
    \begin{enumerate}[label=(\arabic*)]
    \setcounter{enumi}{3}
        \item (Convexity) For all $\alpha \in [0,1]$, $v,w \in \mcL(\Omega)$, $\rho(\alpha v + (1-\alpha) w) \le \alpha \rho(v) + (1-\alpha) \rho(w)$;
    \end{enumerate}
        and coherent if
    \begin{enumerate}[label=(\arabic*)]
    \setcounter{enumi}{4}
        \item (Positive homogeneity) For all $\lambda \ge 0$, $v \in \mcL(\Omega)$, $\rho(\lambda v) = \lambda \rho (v)$.
    \end{enumerate}
\end{defn}

In the following, we introduce some popular kinds of risk measures that will be of interest.

\begin{defn}[Definition 4.112, \cite{Follmer2016}]\label{defn-UBSR}
    A risk measure on $(\Omega,P(\cdot))$ is called a utility-based shortfall risk (UBSR) measure if it can be represented as:
    \[ \SR(v) := \inf\left\{ m \in \mbR : \mbE[\ell(v-m)] \le 0 \right\},\quad \forall v\in \mcL(\Omega), \]
    for some continuous non-decreasing convex loss function $\ell: \mbR \to \mbR$ such that $\ell(0)=0$.\footnote{\cite{Shen2014neuralcomp} employs equivalently $\mbE[\bar{\ell}(v-m)] \le m_0$ using the replacement $\ell(z):=\bar{\ell}(z)-m_0$. We also focus on normalized UBSR.} 
\end{defn}

\begin{exmp}[Expected value]
    When $\ell(x) = x$, the UBSR measure reduces to the expected value, which we refer as the risk-neutral measure. 
\end{exmp}

\begin{exmp}[Entropic risk measure]
    When $\ell(x) = \mathrm{e}^{\beta x}-1$, with $\beta > 0$ representing risk sensitivity, the resulting UBSR measure is the entropic risk measure $\SR(v) = \frac{1}{\beta} \log(\mathbb{E}[\mathrm{e}^{\beta v}])$.
\end{exmp}

\begin{exmp}[Expectile]\label{exmp-expectile}
    Following \cite{Bellini2015}, the expectile is the only coherent UBSR, defined using the loss function \(\ell(x) = \tau x^+ - (1-\tau) x^-\), where \(\tau \in [0,1]\) represents the degree of risk aversion. This measure spans from the essential infimum of the random cost at \(\tau = 0\) to its essential supremum at \(\tau = 1\), passing through the expected value at \(\tau = 0.5\).
\end{exmp}

\begin{defn}[Definition 2.1, \cite{BenTal2007mafi}]\label{defn-OCE}
    A risk measure on $(\Omega,P(\cdot))$ is called an optimized certainty equivalent (OCE) risk measure if it can be represented as
    \begin{align*}
        \OCE(v) := \inf_{\xi\in\mbR}\left\{ \xi + \mbE[\ell(v-\xi)] \right\},\quad \forall v\in\mcL(\Omega),
    \end{align*}
    for some nondecreasing convex loss function $\ell:\mbR\to\mbR$ such that $\ell(0) = 0$ and $1 \in \partial \ell(0)$, where $\partial \ell(0)$ is the subgradient of $\ell$ at $0$.
\end{defn}

\begin{defn}[Definition 3.1, \cite{Acerbi2002jbf}]
    A risk measure on $(\Omega,P(\cdot))$ is called a spectral risk measure associated to a risk spectrum function $\phi:[0,1] \to [0,\infty)$ such that $\int_0^1 \phi(\beta) d\beta = 1$, if it can be represented as 
    \begin{align*}
        M^{\phi}(v) := \int_0^1 \phi(\beta) F_v^{-1}(\beta) d\beta,\quad \forall v\in\mcL(\Omega),
    \end{align*}
    where $F_v$ is the cumulative distribution function of $v$ and $F_v^{-1}(\beta):=\inf\{m \in \mbR: F_v(m) \ge \beta\}$.
\end{defn}

\begin{exmp}[Conditional Value-at-Risk]
    When $\ell(x) = (1-\alpha)^{-1}x^+$ for $\alpha \in (0,1)$, the OCE risk is the conditional Value-at-Risk (CVaR) at level $\alpha$. CVaR is coherent and it is also a spectral risk measure with risk spectrum $\phi(\beta) = (1-\alpha)^{-1}\bm{1}\{\beta \ge \alpha\}$. 
\end{exmp}

\begin{exmp}[Mean-CVaR]
    When $\phi(\beta) = \eta + (1-\eta)(1-\alpha)^{-1}\bm{1}\{\beta \ge \alpha\}$ for some $\eta\in(0,1)$, the spectral risk measure defined as $M^\phi(v) = \eta\mbE[v] + (1-\eta)\CVaR_\alpha(v)$ is referred to as the mean-CVaR risk measure.
\end{exmp}

\subsection{Average Risk-aware MDPs}

We consider a finite MDP defined through the tuple $(\mcX, \mcA, P, c,x_0)$, where $\mcX$ and $\mcA$ are finite state and action spaces, denoting $\mathcal{K} := \mcX \times \mcA$ for short. 
The transition kernel $P: \mcX \times \mcA   \to \mathscr{P}(\mcX)$  specifies the probability $P(y|x, a)$ of transitioning from state $x $ to state $y$ given action $a$. The bounded cost function is defined as $c: \mcX \times \mcA \to [-\bar{C}, \bar{C}]$. For time $t=0,1,\cdots$, the state and action are $x_t$ and $a_t$, governed by a Markov policy $\bm{\pi} = (\pi_0, \pi_1, \dots) $, with each $\pi_t\in\Pi:=\{\pi:\mcX\rightarrow\mathscr{P}(\mcA)\}$, where $\pi_t(\cdot | x_t) $ denotes the probability of choosing $a_t$ given $x_t$. A policy is called deterministic if it assigns a probability of one to a specific action for each state, and is called stationary if $\pi_t \equiv \pi$ for all $t$ for some $\pi\in\Pi$.

In a risk-neutral setting, the infinite horizon average cost MDP problem takes the form:
\[\mbox{(ACMDP)} \quad \bar{J}^* := \inf_{\bm{\pi}\in\Pi^\infty} \limsup_{T\to\infty} \frac{1}{T}  \mbE\left[\sum_{t=0}^T c^{\pi}(X_t)\right],\]
where $X_t$ is the state at stage $t$ and $c^\pi(x) := \sum_{a\in\mcA} \pi(a|x)c(x,a)$.
One seeks to identify a stationary policy that minimizes the long term average expected total cost generated by the MDP, when starting from some initial state $X_0$ and following policy $\bm{\pi}$.

Following \cite{Shen2013sicon}, we consider the risk-aware version of the average cost MDP by replacing $\mbE[\cdot]$ with a class of dynamic risk measures that is specially designed for MDPs.

\begin{defn}
    A risk map $\mcR$ is a function that maps each state $(x,a) \in \mcK$ to a monetary risk measure on the space $(\mcX,P(\cdot|x,a))$. Furthermore, for any $\pi\in\Pi$ we define $\mcR^\pi(v|x):=\sum_{a\in\mcA} \pi(a|x) \mcR(v|x,a)$. To simplify notation, we sometimes write $\mcR_{x,a}(v):=\mcR(v|x,a)$ and $\mcR_x^\pi(v) := \mcR^\pi(v|x)$.
\end{defn}

We first consider a risk-aware $T$-stage total cost problem and define our risk-aware objective as follows:
    $J_T(\bm{\pi}) := c^{\pi_0}(X_0) + \mcR_{X_0}^{\pi_0}(c^{\pi_1}(X_1) + \cdots+ \mcR_{X_{T-1}}^{\pi_{T-1}}(c^{\pi_T}(X_T))\cdots)$.
The infinite horizon average risk-aware MDP problem therefore seeks to find a policy $\bm{\pi}$ that minimizes  :
\[ \mbox{(ARMDP)}\quad  J^* := \inf_{\bm{\pi}\in\Pi^\infty} J_\infty(\bm{\pi}), \]
where $J_\infty(\bm{\pi}) := \limsup_{T\to\infty} \frac{1}{T} J_T(\bm{\pi})$. It is easy to see that ARMDP reduces to ACMDP when $\mcR_{x,a}(v)=\mbE_{x,a}[v]:=\mbE[v(y)]$ with $y\sim P(\cdot|x,a)$.

\begin{remark}
    As argued in \cite{Shen2013sicon}, preserving the Markov property is essential to guarantee stationary optimal policies for infinite horizon objectives \citep{Ruszczynski2006mor, Shen2013sicon}. Therefore, we restrict our attention to Markovian risk measures that depend only on the current state. Readers can refer to \cite{Ruszczynski2010} for a broader framework. 
\end{remark}

\subsection{Average Risk Optimality Equation}\label{sec-AROE}
\cite{Shen2013sicon} establishes several assumptions on the risk maps of an MDP to guarantee the existence and uniqueness of the optimal average risk for ARMDP. Here, we modify and adapt these assumptions to suit our setting of a finite MDP.

\begin{assumption}[Doeblin type condition, Assumption 5.4, \cite{Shen2013sicon}]\label{assump-AROE-Doeblin}
There exists a coherent risk measure $\nu:\mcL(\mcX)\rightarrow \mbR$, and some constant $\bar{\alpha} \in (0,1)$ such that for all $v\ge v' \in \mcL(\mcX)$, we have
$    \min_{(x,a)\in\mcK} \{ \mcR(v|x,a) - \bar{\alpha} \nu(v) - \mcR(v'|x,a) + \bar{\alpha} \nu(v')\} \ge 0$.
\end{assumption}

Assumption \ref{assump-AROE-Doeblin} defines a form of ergodicity property of each state under the risk map. 
In \cite{Shen2013sicon}, $\nu$ is not necessarily required to be coherent, whereas we impose this condition here for simplicity in the subsequent derivations.

From \cite{Shen2013sicon}, if the risk maps satisfy Assumption \ref{assump-AROE-Doeblin}, then there exists an optimal stationary deterministic Markov policy $\bm{\pi}^*$ such that $J^* = J_\infty(\bm{\pi}^*)$. We restate the result as follows.

\begin{thm}[Theorem 5.9, 5.10, \cite{Shen2013sicon}]\label{thm-AROE-generalR}
    Under Assumption \ref{assump-AROE-Doeblin}, there exists a unique $g^* \in \mbR$ and an $h^* \in \mcL(\mcX)$ satisfying the average risk optimality equation (AROE):
    \begin{equation}\label{eq-AROE-generalR}
        g + h(x) = \min_{a\in\mcA} \{ c(x,a) + \mcR(h|x,a)\}.
    \end{equation}
    Moreover, $g^* = J^* = J_\infty(\bm{\pi}^*)$, for the stationary deterministic policy $\pi_t^*(a|x)=\bm{1}\{a=a^*(x)\}$, where $a^*(x)$ minimizes $c(x,a) + \mcR(h^*|x,a)$, and $g^*$ is independent of $x_0$. 
\end{thm}

\begin{remark}\label{remark-Lyapunov}
    In Assumption 5.4 of \cite{Shen2013sicon}, an additional Lyapunov-type condition is introduced, which imposes a growth constraint using a nonnegative weight function $W$. This condition can be dropped in a finite MDP, see Appendix \ref{sec-ProofRQI}.
\end{remark}

\begin{remark}
    Assumption \ref{assump-AROE-Doeblin} is a sufficient condition for the existence of an optimal average risk independent of the initial state and is stronger than the unichain assumption commonly used in risk-neutral average MDPs. 
    It is well-known that for risk-aware MDPs, the unichain assumption alone does not guarantee this independence. For specific risk measures, such as the entropic risk measure, this condition can be relaxed to require only that the Markov chain is irreducible and aperiodic under all stationary policies (see \cite{Cavazos1999mmor}).
    For the three types of risk measures considered in this work, Assumption \ref{assump-AROE-Doeblin} is restricted to an ergodicity condition on the underlying Markov chain, together with a condition on the loss function for UBSR and OCE, and on the risk spectrum for the spectral risk measure (see Theorem \ref{thm-UBSROCEproperties}).
\end{remark}

\section{Model-based Algorithms}\label{sec-ModelBased}
In this section, we propose a risk-aware version of the RVI algorithm to solve the ARMDP problem. Additionally, to lay the foundation for the Q-learning algorithm in the next section, we introduce a risk-aware relative Q-factor iteration algorithm as a generalization of the risk-aware RVI algorithm.

\subsection{Risk-aware Relative Value Iteration}
Following \cite{Abounadi2001sicon}, the risk-neutral RVI algorithm is defined as
\begin{equation}\label{algo-RVI-RN}
    V_{n+1}(x) := \min_{a \in \mcA} \mbE\left[c(x,a) + V_n\right] - f(V_n),\quad \forall x\in \mcX,
\end{equation}
where $V_n \in \mcL(\mcX)$ and $V_0$ is arbitrarily initialized, and $f:\mcL(\mcX)\rightarrow \mbR$ is a function that satisfies conditions discussed below, e.g. $f(v):=v(x_0)$.\footnote{The RVI algorithm in \cite{Bertsekas2007book}, for example, replaces $f(V_n)$ with $f(V_{n+1}):=V_{n+1}(x_0)$. We adopt the formulation from \cite{Abounadi2001sicon} as it is better suited for designing a Q-learning algorithm.}
Using a general function $f(V_n)$ instead of $V(x_0)$ allows the RVI algorithm to eliminate the need for a reference state, making it more flexible and efficient for computation (also see the discussion in \cite{Wan2021icml} for the risk-neutral case). It is known that under the unichain assumption, the risk-neutral RVI algorithm converges to a unique $V^*$, which solves the risk-neutral version of the AROE using $h:=V^*$ and $g^*=f(V^*)$. 

We propose extending the RVI algorithm to the risk-aware setting by replacing the expected value operator with an appropriate risk map. This gives rise to the following risk-aware RVI algorithm: 
    \begin{equation}\label{algo-RVI-generalR}
        V_{n+1}(x) = \mcG(V_n)(x) - f(V_n), \quad \forall x\in \mcX,
    \end{equation}
where $\mcG:\mcL(\mcX) \to \mcL(\mcX)$ is the risk-aware Bellman optimality operator, defined as $\mcG(v)(x) := \min_{a\in\mcA}\mcR_{x,a}(c(x,a) + v)$ for all $x\in\mcX$ and $v\in\mcL(\mcX)$.

To guarantee convergence of algorithm \eqref{algo-RVI-generalR}, we impose the following conditions on $f$.
\begin{assumption}\label{assump-RVI-f} The function $f:\mcL(\mcX) \to \mbR$ satisfies:
    \begin{enumerate}[label=(\arabic*)]
        \item For any $\lambda \in \mbR$ and $v \in \mcL(\mcX)$, $f(\bm{0}) = 0$, $f(v + \lambda) = f(v) + \lambda$.
        \item $f$ is Lipschitz, i.e., $\exists \tilde{L} \ge 0$, such that $\norm{f(v) - f(w)}_\infty \le \tilde{L}\norm{v-w}_\infty,\, \forall v, w \in \mcL(\mcX)$.
    \end{enumerate}
\end{assumption}
Assumption \ref{assump-RVI-f} is equivalent to imposing that $f$ is translation invariant and is naturally satisfied by $f(v) := v(x_0)$. 
Such extension for the RVI seems to be first proposed in \cite{Abounadi2001sicon} for the risk-neutral case, although the authors did not include proofs of their validity. 

The following theorem confirms that convergence of RVI remains valid in the risk-aware setting.
\begin{thm}\label{thm-RVI-generalR}
    Under assumptions \ref{assump-AROE-Doeblin} and \ref{assump-RVI-f},
    the risk-aware RVI algorithm \eqref{algo-RVI-generalR}
     converges to a unique fixed point $V^*$, which identifies, using $h^*(x):=V^*(x)$ and $g^*:=f(V^*)$, a solution to the AROE \eqref{eq-AROE-generalR}. 
\end{thm}

Clearly, the risk-aware RVI algorithm \eqref{algo-RVI-generalR} reduces to 
RVI algorithm \eqref{algo-RVI-RN} in the risk-neutral setting. 

\subsection{Risk-aware Relative Q-factor Iteration}
The risk-aware RVI algorithm \eqref{algo-RVI-generalR} suggests that, when letting $Q_{n+1}(x,a) := \mcR_{x,a}(c(x,a) + V_n)-f(V_n)$, \eqref{algo-RVI-generalR} can be reformulated as the following risk-aware relative Q-factor iteration (RQI):
\begin{align*}
    Q_{n+1}(x,a) := \mcR_{x,a}\left(c(x,a) + \min_{a'\in\mcA} Q_n(\cdot,a')\right)- f\left(\min_{a'\in \mcA}Q_n(\cdot,a')\right),\quad \forall (x,a) \in \mcK,
\end{align*}
where $Q_n \in \mcL(\mcK)$ and $Q_0$ is arbitrarily initialized. As suggested in \cite{Abounadi2001sicon} for the risk-neutral case, this can be more generally defined as: 
\begin{equation}\label{algo-RQI-generalR}
\begin{aligned}
    Q_{n+1}(x,a) = \mcH(Q_n)(x,a) - f(Q_n), \quad \forall (x,a) \in \mcK,
\end{aligned}
\end{equation}
where $\mcH: \mcL(\mcK) \to \mcL(\mcK)$ is the risk-aware Bellman optimality operator for Q-factors, defined as $\mcH(q)(x,a):= \mcR_{x,a}(c(x,a) + \min_{a'\in\mcA} q(\cdot,a'))$, for all $(x,a) \in \mcK$ and $q\in\mcL(\mcK)$.
With a slight abuse of notation, here we define $f:\mcL(\mcK) \to \mbR$ and impose the following assumptions.
\begin{assumption}\label{assump-RQI-f}
    The function $f:\mcL(\mcK) \to \mbR$ satisfies:
    \begin{enumerate}[label=(\arabic*)]
        \item For any $\lambda \in \mbR$ and $q \in \mcL(\mcK)$, $f(\bm{0}) = 0$, $f(q + \lambda) = f(q) + \lambda$.
        \item $f$ is Lipschitz, i.e., $\exists \tilde{L} \ge 0$ such that $\norm{f(p) - f(q)}_\infty \le \tilde{L}\norm{p-q}_\infty$, $\forall p, q \in \mcL(\mcK)$.
    \end{enumerate}
\end{assumption}
Common choices for $f$ can be 
$f(q) = q(x_0,a_0)$, $f(q) = \min_a q(x_0,a)$, $f(q)=\frac{1}{|\mcX||\mcA|}\sum_{x,a}q(x,a)$. 
Similar to Theorem \ref{thm-RVI-generalR}, we have the following convergence and optimality result for the risk-aware RQI algorithm.

\begin{thm}\label{thm-RQI-generalR}
    Under assumptions \ref{assump-AROE-Doeblin} and \ref{assump-RQI-f}, the risk-aware RQI algorithm \eqref{algo-RQI-generalR}
    converges to a unique fixed point $Q^*$, which identifies, using $h^*(x):=\min_{a\in\mcA}Q^*(x,a)$ and $g^*:=f(Q^*)$, a solution to the AROE \eqref{eq-AROE-generalR}.
\end{thm}

Theorem \ref{thm-RQI-generalR} also suggests that a solution to the AROE \eqref{eq-AROE-generalR} can be identified by solving the following average risk optimality equation based on the Q-factor:
\begin{equation}\label{eq-RQI-AROE}
    q(x,a) =\mcR_{x,a}\left(c(x,a) +  \min_{a'\in\mcA} q(\cdot,a')\right) - f(q), \quad \forall(x,a)\in\mcK, q \in \mcL(\mcK).
\end{equation}

\section{Model-free Algorithms}\label{sec-ModelFree}
In this section, we propose model-free Q-learning algorithms for solving the ARMDP. We begin by showing that almost sure convergence can be achieved when an unbiased estimator of the risk-aware Bellman optimality operator is available. In particular, we describe how such an estimator can be constructed using the multi-level Monte Carlo method. In addition, we introduce a Q-learning algorithm tailored to UBSR that avoids the need for an estimator.

\subsection{Risk-aware RVI Q-learning}
Motivated by the risk-aware RQI algorithm \eqref{algo-RQI-generalR}, we can propose the following model-free risk-aware RVI Q-learning algorithm:
\begin{equation}\label{algo-RVIQ-generalR}
    Q_{n+1}(x,a) = Q_n(x,a) + \gamma(n) \left(\hat{\mcH}(Q_n)(x,a) - f(Q_n) - Q_n(x,a)\right),\quad (x,a) \in \mcK,
\end{equation}
where $\hat{\mcH}$ is an estimator for the risk-aware Bellman optimality operator $\mcH$ and $\gamma(n)$ is some step size. 
We construct $\hat{\mcH}(q)$ as an estimator of $\mcH(q)$ satisfying the following assumption.
\begin{assumption}\label{assump-unbiased}
    The estimator $\hat{\mcH}$ is unbiased and has controllable variance: $\mbE[\hat{\mcH}(q)] = \mcH(q)$ and there exists a $C > 0$ such that $\Var[\hat{\mcH}(q)(x,a)] \le C (1 + \norm{q}_\infty^2)$, $\forall (x,a) \in \mcK, q\in\mcL(\mcK)$.
\end{assumption}

To guarantee the convergence for the risk-aware RVI Q-learning algorithm \eqref{algo-RVIQ-generalR}, we require that the risk map satisfies an assumption called ``asymptotic coherence" and the function $f$ is homogeneous. We also impose the Robbins-Monro condition on the step size $\gamma(n)$.

\begin{assumption}[Asymptotic coherence]\label{assump-asymptcoherent}
    The risk map $\mcR$ is asymptotically coherent, i.e., there exists a risk map $\mcR^\infty$ such that for all $(x,a)\in\mcK$, we have that $\lim_{s\rightarrow \infty} \frac{1}{s}\mcR_{x,a}(s v)=\mcR_{x,a}^\infty(v)$ for all $v\in\mcL(\mcX)$
    and uniformly on all compact subsets of $\mcL(\mcX)$. 
\end{assumption}

\begin{assumption}\label{assump-fhomogeneous}
    The function $f$ is homogeneous, i.e., $f(\lambda v) = \lambda f(v)$, $\forall \lambda\in\mbR$, $v\in\mcL(\mcK)$.
\end{assumption}

\begin{assumption}\label{assump-RobbinsMonro}
    The step size $\{\gamma(n)\}_{n=0}^\infty$ satisfies $\sum_{n=0}^\infty \gamma(n) = \infty$ and $\sum_{n=0}^\infty \gamma(n)^2 < \infty$.
\end{assumption}

We then have the following convergence result.
\begin{thm}\label{thm-RVIQ-generalR}
    Under assumptions \ref{assump-AROE-Doeblin}, \ref{assump-RQI-f}, \ref{assump-unbiased}, \ref{assump-asymptcoherent}, \ref{assump-fhomogeneous}, \ref{assump-RobbinsMonro}, then almost surely, $Q_n$ converges to some $Q^*$, and $h^*(x):=\min_{a\in\mcA}Q^*(x,a)$, $g^*:=f(Q^*)$ identify a solution to the AROE \eqref{eq-AROE-generalR}. The greedy policy, $\pi_n(a|x):=\bm{1}\{a=a_n^*(x)\}$ with $a_n^*(x)\in \arg\min_{a\in\mcA} Q_n(x,a)$, also converges almost surely to an optimal stationary deterministic policy of ARMDP.
\end{thm}

If the risk map $\mcR$ is coherent, Assumption \ref{assump-asymptcoherent} is automatically satisfied. This assumption is made for technical reasons, as the convergence proof of the Q-learning algorithm for the average MDP relies on ODE-based stochastic approximation (see \cite{Abounadi2001sicon} and \cite{Wan2024arXiv} for a recent review). This approach requires the limit in Assumption \ref{assump-asymptcoherent} to exist for analyzing the almost sure boundedness of the iteration sequence. In the next subsection, we demonstrate that this assumption can be achieved for many risk maps that are not necessarily coherent.

\subsection{Construction of an Unbiased Estimator $\hat{\mcH}$} 
In this subsection, we present an estimator \(\hat{\mathcal{H}}\) that satisfies Assumption \ref{assump-unbiased} for specific risk maps using the multi-level Monte Carlo (MLMC) method, an approach for unbiased statistical estimation using stochastic simulation \citep{Blanchet2015wsc, Blanchet2019arXiv, Liu2022icml, WangSB2023aistats, WangY2023icml}. 
We first impose the following assumption on $\mcR$, which enables the possibility of estimating a random variable using its empirical distribution.

\begin{assumption}[Hölder continuity]\label{assump-WassLipschitz}
    There exists an $\mfL > 0$ such that for all $v, w \in \mcL(\mcX)$, we have $|\mcR_{x,a}(v) - \mcR_{x,a}(w)| \le \mfL d_W(\mu_v, \mu_w)$, $\forall (x,a)\in\mcK$, where $\mu_v$, $\mu_w$ are the probability distributions of $v$ and $w$ on $(\mcX,P(\cdot|x,a))$ and $d_W(\cdot,\cdot)$ is the 1-Wasserstein distance between two distributions.
\end{assumption}

We first generate $N$ according to a geometric distribution with parameter $r \in (0,1)$. Then, for each $(x,a) \in \mcK$, we take action $a$ at state $x$ for $2^{N+1}$ times and observe the i.i.d. transitions $\{x_i'\}_{i=1}^{2^{N+1}}$. 
These \(2^{N+1}\) samples are then divided into two groups: samples with odd indices and samples with even indices. We calculate the empirical distribution of \(x'\) using the even-index samples, odd-index samples, all the samples, and the first sample: 
$\hat{P}_{N+1}^{E}(y|x,a) := \frac{1}{2^N}\sum_{i=1}^{2^N} \bm{1}\{x_{2i}'=y\}$, $\hat{P}_{N+1}^{O}(y|x,a) := \frac{1}{2^N}\sum_{i=1}^{2^N} \bm{1}\{x_{2i-1}'=y\}$, $\hat{P}_{N+1}(y|x,a) := \frac{1}{2^{N+1}}\sum_{i=1}^{2^{N+1}} \bm{1}\{x_{i}'=y\}$, $\hat{P}_{N+1}^1(y|x,a) := \bm{1}\{x_1'=y\}$.
For notation simplicity, we denote the resulting empirical transition kernels as $\hat{P}_{N+1}^{E}$, $\hat{P}_{N+1}^{O}$, $\hat{P}_{N+1}$ and $\hat{P}_{N+1}^{1}$, respectively. Then, we use these estimated transition kernels as nominal kernels to calculate $\mcH$. Namely, $\mcH_{\hat{P}_{N}}$ is the Bellman optimality operator under the empirical transition kernel $\hat{P}_{N}$. The multi-level estimator of $\mcH$ is then defined as
\begin{equation}\label{eq-mcHQ}
    \hat{\mcH}(q) := \mcH_{\hat{P}_{N+1}^{1}}(q) + \frac{1}{p_N}\left\{ \mcH_{\hat{P}_{N+1}}(q) 
    - \frac{1}{2}\left(\mcH_{\hat{P}_{N+1}^{E}}(q) + \mcH_{\hat{P}_{N+1}^{O}}(q)\right) \right\},\quad \forall q\in\mcL(\mcK),
\end{equation}
where $p_N := r(1-r)^N$.

We present the following result on the unbiasedness and controllable variance for risk maps satisfying Assumption \ref{assump-WassLipschitz}.
\begin{thm}\label{thm-unbiasedriskmap}
    Assumption \ref{assump-unbiased} holds if the risk map satisfies Assumption \ref{assump-WassLipschitz} and $r\in(0,1/2)$.
\end{thm}

As shown in Section 3 of \cite{Prashanth2022jmlr}, several popular risk measures, including UBSR, OCE and spectral risk measures, satisfy Assumption \ref{assump-WassLipschitz} with proper parametrization. Below, we demonstrate that under suitable conditions, the average risk-aware MDP that incorporates UBSR, OCE, or spectral risk measure satisfies all the assumptions required for Theorem \ref{thm-RVIQ-generalR} to apply.

\begin{assumption}[Strong ergodicity]\label{assump-ergodicity}
    Under any stationary policy, the resulting Markov chain is irreducible and there exists a state $\bar{x}\in\mcX$ such that $P(\bar{x}|x,a) > 0$ , $\forall (x,a) \in \mcK$.
\end{assumption}

\begin{assumption}[Bounded slope]\label{assump-boundedslope}
    The loss function $\ell(x)$ is strictly increasing on $\mbR$ and there exist $L_1,\epsilon_1>0$ such that $0<\epsilon_1\le \frac{\ell(x) - \ell(y)}{x-y} \le L_1$, $\forall x\neq y\in \mbR$.
\end{assumption}

\begin{thm}\label{thm-UBSROCEproperties}
    Under Assumption \ref{assump-ergodicity}, if the risk map employs a UBSR or OCE satisfying Assumption \ref{assump-boundedslope}, or a spectral risk measure with $0 < \epsilon_2 \le \phi(\cdot) \le L_2 < \infty$,
    then assumptions \ref{assump-AROE-Doeblin}, \ref{assump-asymptcoherent}, and \ref{assump-WassLipschitz} hold. Consequently, Theorem \ref{thm-RVIQ-generalR} applies.
\end{thm}

We note that CVaR does not satisfy the condition in Theorem \ref{thm-UBSROCEproperties}. However, the mean-CVaR risk measure, which mixes expectation and CVaR, does and hence Theorem \ref{thm-RVIQ-generalR} applies for mean-CVaR. Also, although our definitions of UBSR and OCE assume a convex loss function, Theorem \ref{thm-UBSROCEproperties} holds more generally for loss functions that are convex (concave) for \(x > 0\) and concave (convex) for \(x < 0\), reflecting different risk attitudes toward gains and losses (see Appendix \ref{app:thmUBSROCESRM}). Finally, the entropic risk measure does not satisfy Assumption \ref{assump-boundedslope}, but it still meets Assumption \ref{assump-AROE-Doeblin} (Proposition 5.7, \cite{Shen2013sicon}). \cite{Borkar2002mor} proposed a Q-learning algorithm for average risk-aware MDPs with an entropic risk measure, which is derived from the multiplicative Poisson equation and does not rely on MLMC. For further details, readers may refer to \cite{Borkar2002mor}.

\subsection{An Off-policy Q-learning Algorithm for UBSR Measures}\label{sec-UBSRQ-learning}
An important practical concern of our MLMC Q-learning algorithm is the necessity for a resampling procedure for each $(x,a)$ pair, which prevents the algorithm from being adapted for off-policy learning. This can be addressed when the risk map employs a UBSR measure using an approach proposed in \cite{Shen2014neuralcomp} for risk-aware discounted MDPs. Namely, Proposition 4.113 in \cite{Follmer2016} establishes that for any $v\in\mcL(\mcX)$, the risk map $\SR_{x,a}(v)$ is the unique solution of $\mbE_{x,a}[\ell(v-\SR_{x,a}(v))] = 0$. This implies that the AROE \eqref{eq-RQI-AROE} can be equivalently rewritten as:
\begin{equation*}
    \mbE\left[\ell \left( c(x,a) + \min_{a'\in\mcA} q(\cdot,a') - f(q) - q(x,a) \right)\right]= 0,\quad \forall (x,a) \in \mcK.
\end{equation*}
This motivates the following off-policy algorithm that seeks to identify the root of this AROE using stochastic approximation (see \cite{Borkar2008book}). Specifically, given any sequence $\{(x_n,a_n,x_n')\}$, with $x_n'\sim P(\cdot|x,a)$, the asynchronous UBSR-based Q-learning consists in applying the updates:
\begin{equation}\label{algo-RVIQ-UBSR-Asynchronous}
    Q_{n+1}(x_n,a_n) = Q_n(x_n,a_n) + \gamma(n) \ell \Big( c(x_n,a_n) +\min_{a' \in \mcA} Q_n(x_n',a') - f(Q_n) - Q_n(x_n,a_n)  \Big).
\end{equation}
where each subsequence $\{\gamma(n)\}_{n:(x_n,a_n)=(x,a)}$, indexed by $(x,a)\in\mcK$, satisfies Assumption \ref{assump-RobbinsMonro}. The synchronous algorithm can also be derived if all $(x,a)$ pairs are updated within one iteration.

It shall be noticed that the theoretical convergence of this algorithm remains an open question. Drawing from the ODE analysis used in the risk-neutral case, the corresponding ODE for \eqref{algo-RVIQ-UBSR-Asynchronous} involves a high-dimensional, nonlinear system that is difficult to analyze for stability. Additional discussion on this problem can be found in Appendix \ref{sec-AlternativeRVIQUBSR}.

\section{Experiments}\label{sec-Experiments}
In this section, we provide numerical experiments confirming the convergence of our MLMC-based Q-learning algorithm (MLMC Q-learning), comparisons to the off-policy Q-learning algorithm for UBSR (UBSR Q-learning), and apply our algorithm (see pseudo-codes in Appendix \ref{sec-Pseudocode}) to real-life problems to showcase its potential. Further details and experimental investigations are also presented in Appendix \ref{sec-AdditionalExperiments}, namely regarding the sensitivity of MLMC Q-learning to $r$, the convergence of UBSR Q-learning, and the effect of risk-awareness in long term performance of policies.

\subsection{Convergence of MLMC Q-learning}\label{sec-Experiments-ConvergenceRVIQ}
We begin by validating the convergence of the risk-aware RVI Q-learning algorithm \eqref{algo-RVIQ-generalR} using a randomly generated MDP with 10 states and 5 actions per state. The nominal transition kernel $P$ is generated from a uniform distribution over $[0,1]$ and subsequently normalized. The cost function is sampled from a normal distribution $\mathcal{N}(1,1)$. We choose $\gamma(n) := (1/(n+1))^{2/3}$ and $f(q) := \frac{1}{|\mcX||\mcA|} \sum_{x,a} q(x,a)$. 
Due to space limit, we only show the convergence results for two special cases of UBSR and OCE: the expectile with $\tau = 0.75$ and OCE with loss function $\ell(x):=\gamma_1 x^+ - \gamma_2 x^-$ where $\gamma_1 = 2$ and $\gamma_2 = 0.5$. 

\begin{figure}[ht!]
    \centering
    \subfigure{
        \includegraphics[width=0.40\textwidth]{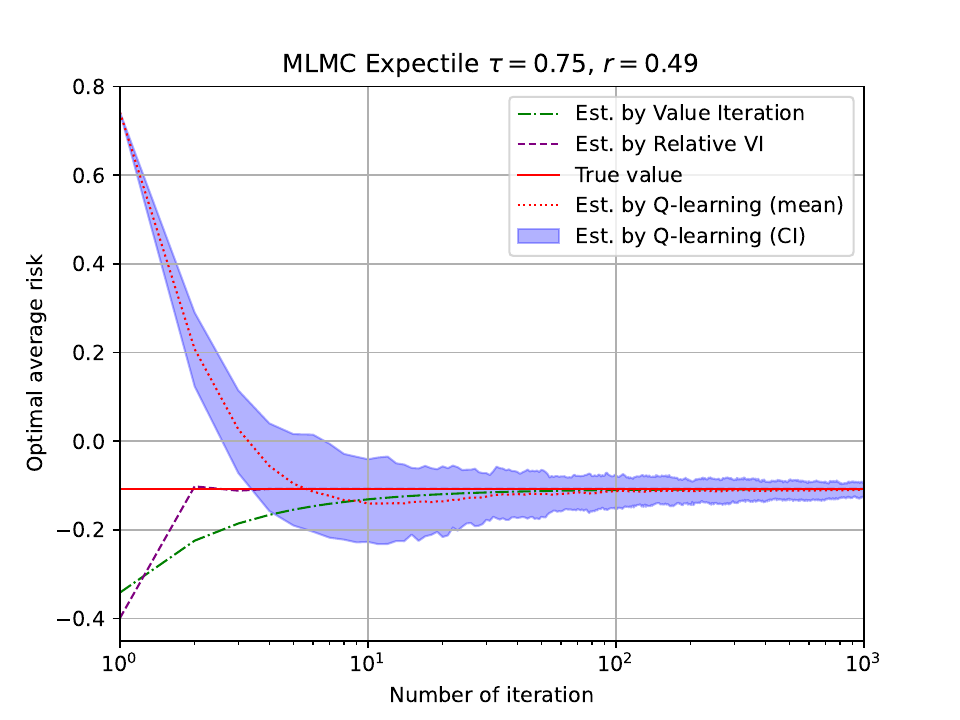}
    }
    \subfigure{
        \includegraphics[width=0.40\textwidth]{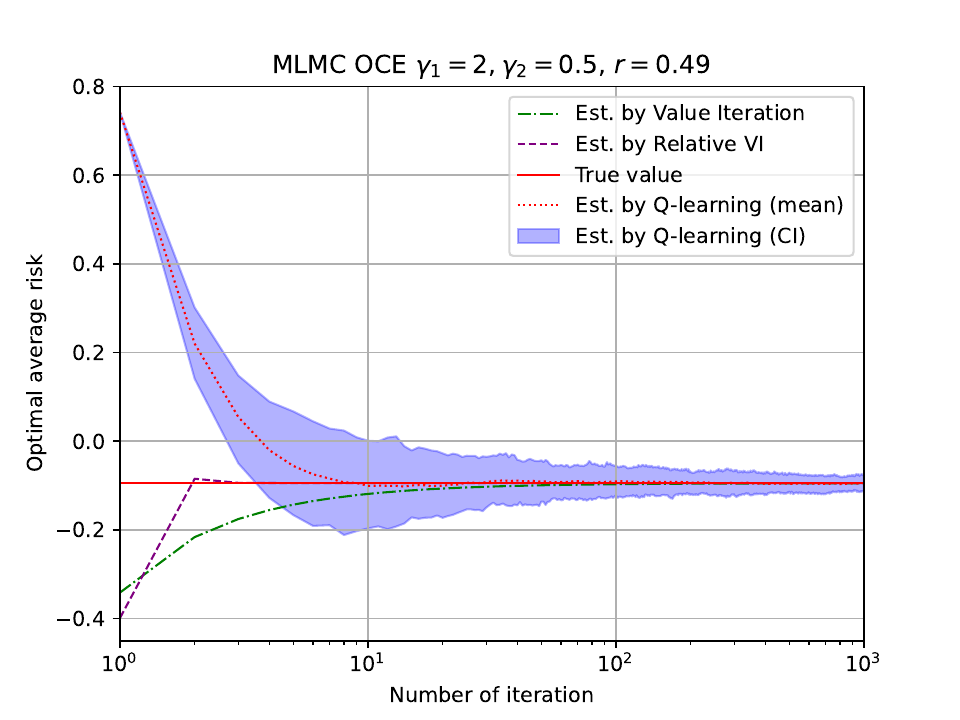}
    }
    \caption{Convergence experiments for risk-aware RVI \eqref{algo-RVI-generalR} and MLMC Q-learning \eqref{algo-RVIQ-generalR}.}
    \label{figure:UBSROCE}
\end{figure}

We run the MLMC Q-learning algorithm 100 times independently with $r=0.49$ and plot the mean value of $f(Q_n)$ in Figure \ref{figure:UBSROCE}, with the 95th and 5th percentiles as the confidence interval (CI). For comparison, trajectories from value iteration and the risk-aware RVI algorithm \eqref{algo-RVI-generalR} are shown together with the true optimal risk (via value iteration). It is evident that the MLMC Q-learning algorithm converges to the true optimal average risk almost surely. As a model-based approach, the risk-aware RVI achieves convergence to the optimum at a significantly faster rate.

It is worth noting that selecting \( r \in (0,1/2) \) does not ensure finite sample guarantees based on Theorem \ref{thm-unbiasedriskmap}, as each iteration requires an average of infinitely many samples when \( r \leq 1/2 \).
However, for some $r \in (1/2,3/4)$, both asymptotic and finite sample guarantees may still be achieved as observed empirically in additional experiments  presented in Appendix \ref{sec-StatisticalMLMC}. These empirical findings are coherent with the guarantees identified in \cite{WangSB2023aistats} for a special class of distributionally robust discounted MDPs. 

\subsection{Comparisons of MLMC and UBSR Q-learning Algorithms}

\begin{figure}[h]
    \centering
    \subfigure{
        \includegraphics[width=0.3\linewidth]{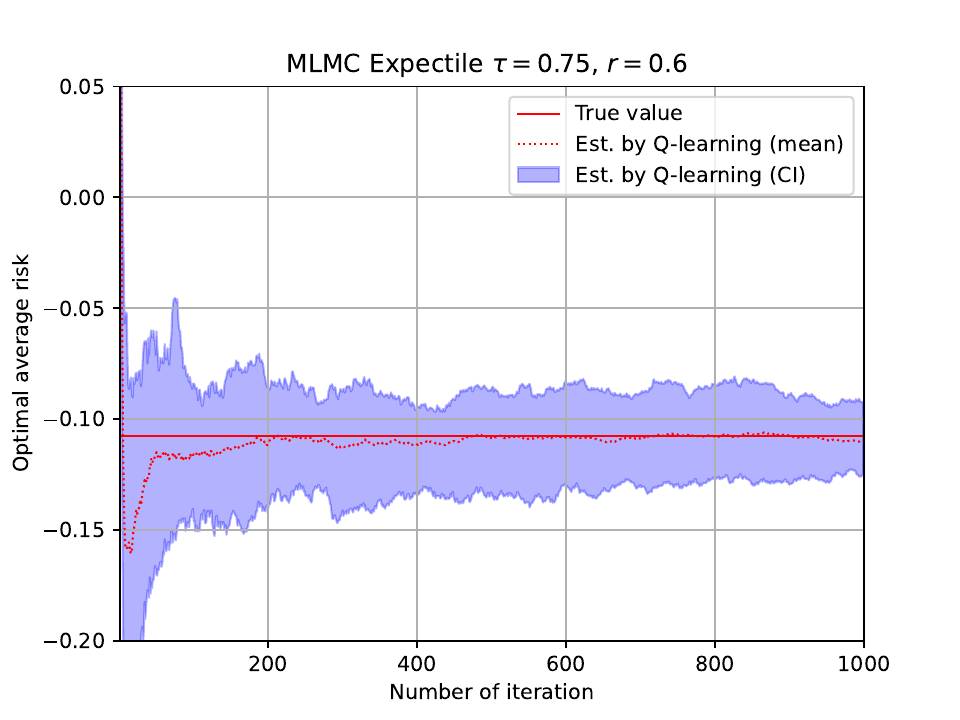}
    }
    \hfill
    \subfigure{
        \includegraphics[width=0.3\linewidth]{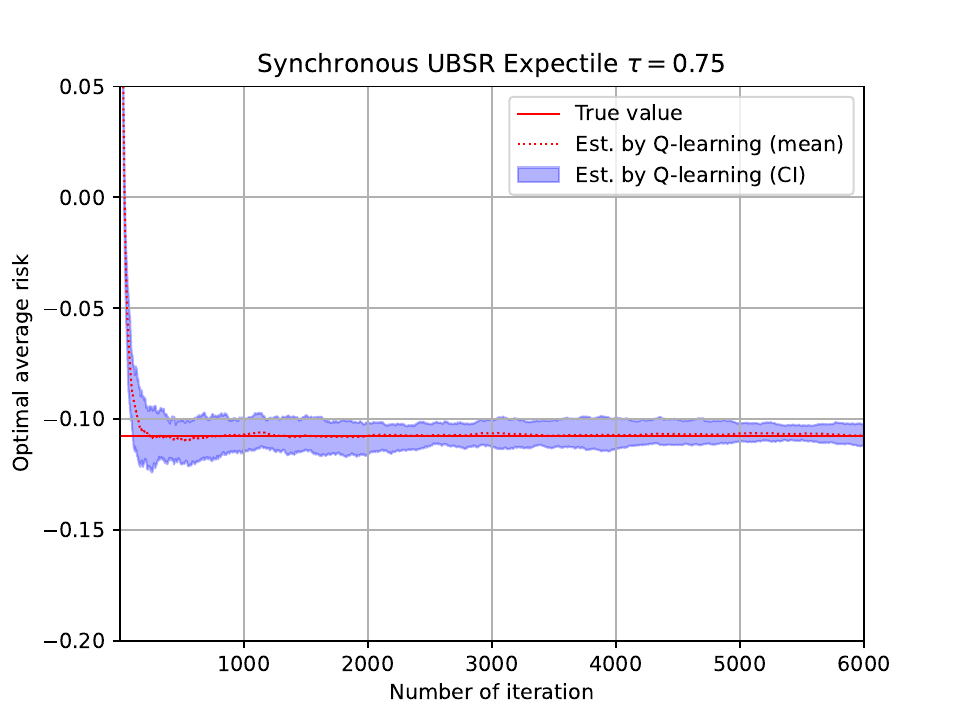}
    }
    \hfill
    \subfigure{
        \includegraphics[width=0.3\linewidth]{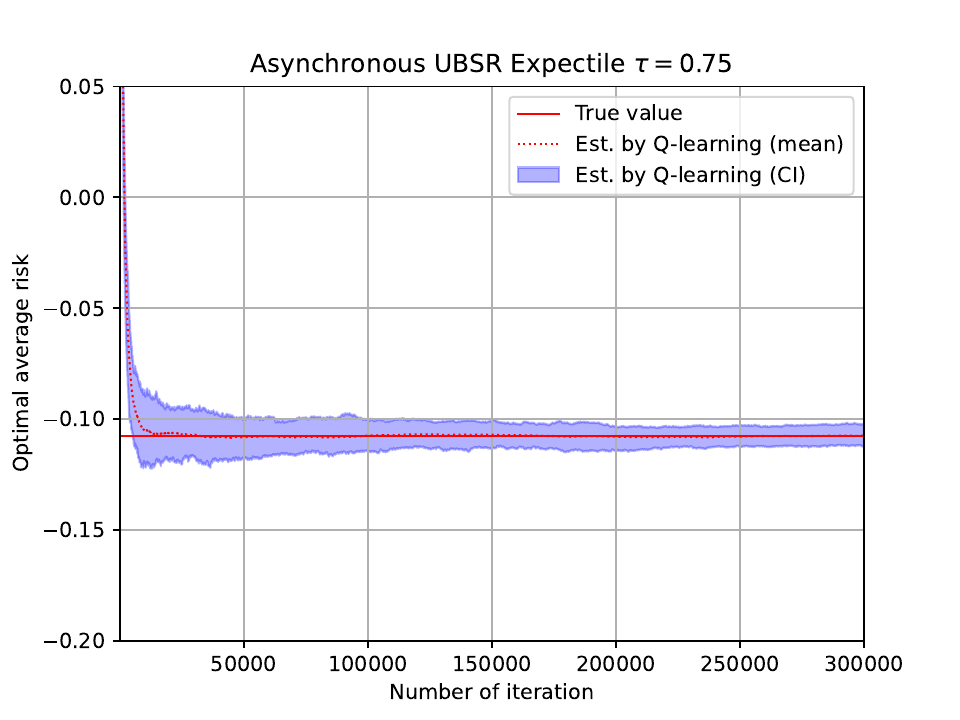}
    }
    \caption{Comparison of MLMC and UBSR Q-learning with equivalent number of samples.}
    \label{figure:UBSRMLMCComparison}
\end{figure}

Figure \ref{figure:UBSRMLMCComparison} presents the convergence  of the synchronous and asynchronous UBSR Q-learning algorithms for the expectile in the same setting outlined in Section \ref{sec-Experiments-ConvergenceRVIQ}. For comparison, we also include the results of the MLMC Q-learning algorithm \eqref{algo-RVIQ-generalR}) with \( r = 0.6 \), which uses an expected 6,000 samples per state-action pair over 1,000 iterations, corresponding to 300,000 iterations of the asynchronous UBSR Q-learning algorithm \eqref{algo-RVIQ-UBSR-Asynchronous} and 6,000 iterations for the synchronous version.

Further experiments with different loss functions along with a detailed discussion and comparison to the MLMC Q-learning algorithm \eqref{algo-RVIQ-generalR} are provided in Appendix \ref{sec-AlternativeRVIQUBSR}. 
While the MLMC Q-learning algorithm offers provable convergence to optimality, the two UBSR Q-learning algorithms \eqref{algo-RVIQ-UBSR-Asynchronous} demonstrate both faster convergence and lower variance compared to the MLMC Q-learning algorithm \eqref{algo-RVIQ-generalR}.
Investigating the almost sure convergence and optimality of this algorithm remains an interesting direction for future research.

\subsection{Applications}
To illustrate the practicality of our risk-aware algorithms, we tested them on three popular average-cost MDP problems: machine replacement (MR), water reservoir management (WR), and inventory management (IM) (e.g. \cite{Puterman1994book}, \cite{Hernandez1989book}). Each problem is evaluated under four risk measures: expectile (EX, $\tau = 0.9$), OCE ($\gamma_1 = 2$, $\gamma_2 = 0.5$), mean-CVaR ($\eta = 0.1, \alpha = 0.2$) and risk-neutral (RN). The optimal policies for each risk measure are obtained using the risk-aware RVI algorithm.
Experimental details are provided in Appendix \ref{sec-ApplicationSetups}.

Table \ref{table-1} reports the optimal average risk for the three applications across the four risk measures, as well as the average risk of the four risk-aware policies when evaluated under the expectile risk measure with $\tau=0.9$, i.e., the average risk computed under the expectile risk measure using the optimal policies derived for the other risk measures. 
It can be observed that the policies produced by our algorithms successfully attain their respective optimal average risks under difference application settings.
Appendix \ref{sec-RiskAnalysisExpectile} explores risk differences across $\tau$ values, showcasing UBSR’s flexibility in risk preference design. These results confirm our theory, proving the effectiveness of our algorithms in computing optimal risk-aware policies tailored to an agent's risk preferences.

\begin{table}[htbp]
  \centering
  \caption{Average risk under different risk measures for three experimental setups: machine replacement (MR), water reservoir management (WR), and inventory management (IM)}
 \resizebox{0.8\textwidth}{!}{
    \begin{tabular}{ccccccc}
    \toprule
    \multirow{2}[4]{*}{\textbf{Risk Measures}} & \multicolumn{3}{c}{\textbf{Optimal Risk}} & \multicolumn{3}{c}{\textbf{Expectile Risk}} \\
\cmidrule{2-7}          & \textbf{MR} & \textbf{WR} & \textbf{IM} & \textbf{MR} & \textbf{WR} & \textbf{IM} \\
    \midrule
    EX    & 68.7499 & 20.2541 & 24.8694 & \textbf{68.7499} & \textbf{20.2541} & \textbf{24.8694} \\
    OCE   & 63.9291 & 14.1389 & 23.7330 & 68.9323 & 20.3239 & 25.2908 \\
    Mean-CVaR & 59.5244 & 9.9319 & 22.8955 & 69.3343 & 20.6413 & 26.5721 \\
    RN    & 54.4233 & 7.6174 & 20.1345 & 69.9359 & 20.6413 & 28.1901 \\
    \bottomrule
    \end{tabular}%
    }
  \label{table-1}%
\end{table}%

\section{Conclusion and Future Research}\label{sec-Conclusion}
In this paper, we introduced the first risk-aware RVI algorithm and two novel model-free risk-aware RVI Q-learning algorithms for average-cost MDPs. MLMC Q-learning can be applied with a more general class of risk measures, while requiring access to repeated samples of transitions from a given state-action pair. UBSR Q-learning more closely aligns with the traditional setting of Q-learning yet is dedicated to the class of UBSR measures.

Several research directions are worth exploring. First, we conjecture that the strong ergodicity Assumption \ref{assump-ergodicity} could be weakened. Second, establishing finite-sample guarantees for MLMC Q-learning remains an open problem. It would be interesting to examine whether the conditions in \cite{WangSB2023aistats} for a specific class of distributionally robust discounted MDPs and the variance reduction techniques in \cite{WangSB2024jmlr} can be extended to our framework. Third, the almost sure convergence of UBSR Q-learning should be addressed. 
Finally, exploring the applicability to large-scale problems and designing other types of reinforcement learning algorithms for average risk-aware reinforcement learning constitute important directions for future research.

\section*{Acknowledgments}
The authors are grateful to the anonymous reviewers for their valuable feedback and insightful
comments. Erick Delage was partially supported by the Canadian Natural Sciences and Engineering Research Council [Grant RGPIN-2022-05261] and by the Canada Research Chair program [950-230057]. We are also thankful to Esther Derman, Marek Petrik and Xian Chen for valuable discussions on related topics.

\bibliography{reference}

\newpage
\appendix
{\Large \textbf{Appendices}}
\section{Algorithm Pseudo-codes}\label{sec-Pseudocode}
\begin{algorithm}[htbp]
    \caption{Risk-aware Relative Value Iteration}
    \begin{algorithmic}[1]
        \STATE \textbf{Input:} $V_0$, $f$, $T$, $n\leftarrow 0$;
        \WHILE{$n < T$}
        \FOR{all $x\in\mcX$}
        \STATE $V_{n+1}(x) \leftarrow \mcG(V_n)(x) - f(V_n)$, where $\mcG$ defined in \eqref{algo-RVI-generalR};
        \ENDFOR
        \STATE $n\leftarrow n+1$;
        \ENDWHILE
    \end{algorithmic}
\end{algorithm}

\begin{algorithm}[htbp]
    \caption{Risk-aware Relative Q-factor Iteration}
    \begin{algorithmic}[1]
        \STATE \textbf{Input:} $Q_0$, $f$, $T$, $n\leftarrow 0$;
        \WHILE{$n < T$}
        \FOR{all $x\in\mcX$, $a\in\mcA$}
        \STATE $Q_{n+1}(x,a) \leftarrow \mcH(Q_n)(x,a) - f(Q_n)$, where $\mcH$ defined in \eqref{algo-RQI-generalR};
        \ENDFOR
        \STATE $n\leftarrow n+1$;
        \ENDWHILE
    \end{algorithmic}
\end{algorithm}

\begin{algorithm}[htbp]
    \caption{Risk-aware RVI Q-learning with MLMC (MLMC Q-learning)}
    \begin{algorithmic}[1]
        \STATE \textbf{Input:} $Q_0$, $f$, $\gamma(n)$, $r\in(0,1)$, $T$, $n\leftarrow 0$;
        \WHILE{$n < T$}
        \FOR{all $x\in\mcX$, $a\in\mcA$}
        \STATE Sample $N\sim\mathrm{Geo}(r)$;
        \STATE Independently draw $2^{N+1}$ samples $x_i'\sim P(\cdot|x,a)$;
        \STATE $Q_{n+1}(x,a) \leftarrow Q_n(x,a) + \gamma(n)\left(\hat{\mcH}(Q_n)(x,a) - f(Q_n) - Q_n(x,a) \right)$, where $\hat{\mcH}$ is defined in \eqref{eq-mcHQ};
        \ENDFOR
        \STATE $n\leftarrow n+1$;
        \ENDWHILE
    \end{algorithmic}
\end{algorithm}

\begin{algorithm}[!htbp]
    \caption{Off-policy (asynchronous) RVI Q-learning for UBSR (A-UBSR Q-learning)}
    \begin{algorithmic}[1]
        \STATE \textbf{Input:} $Q_0$, $\ell$, $f$, $\gamma(n)$, $T$, $n\leftarrow 0$;
        \WHILE{$n < T$}
        \STATE Observe one transition $(x,a,x')$;
        \STATE $Q_{n+1}(x,a) \leftarrow Q_n(x,a) + \gamma(n) \ell\left(c(x,a) + \min_{a'\in\mcA} Q_n(x',a') - f(Q_n) - Q_n(x,a) \right)$;
        \STATE $n\leftarrow n+1$;
        \ENDWHILE
    \end{algorithmic}
\end{algorithm}

\begin{algorithm}[!htbp]
    \caption{Synchronous RVI Q-learning for UBSR (S-UBSR Q-learning)}
    \begin{algorithmic}[1]
        \STATE \textbf{Input:} $Q_0$, $\ell$, $f$, $\gamma(n)$, $T$, $n\leftarrow 0$;
        \WHILE{$n < T$}
        \FOR{all $x\in\mcX$, $a\in\mcA$}
        \STATE Observe one sample $x'$;
        \STATE $Q_{n+1}(x,a) \leftarrow Q_n(x,a) + \gamma(n) \ell\left(c(x,a) + \min_{a'\in\mcA} Q_n(x',a') - f(Q_n) - Q_n(x,a) \right)$;
        \ENDFOR
        \STATE $n\leftarrow n+1$;
        \ENDWHILE
    \end{algorithmic}
\end{algorithm}

\newpage

\section{Proofs}\label{sec-Proof}

\subsection{Proof of Theorem \ref{thm-RVI-generalR} and \ref{thm-RQI-generalR}}\label{sec-ProofRQI}
From the definition of the risk-aware RQI algorithm \eqref{algo-RQI-generalR}, it is evident that the risk-aware RVI algorithm \eqref{algo-RVI-generalR} can be considered a special case of RQI by defining $V(x) := \min_{a\in\mcA} Q(x,a)$. Consequently, if the risk-aware RQI algorithm converges, then it follows that the risk-aware RVI algorithm also converges.

In order to study the Q-factor iteration, we make use of an augmented risk map $\widetilde{\mcR}$, 
on the $\mcK$ outcome space, using 
\begin{equation}\label{eq-augmentedriskmap}
    \widetilde{\mcR}_{x,a}(q) := \mcR\left(\min_{a'\in\mcA} q(\cdot,a') \Big| x,a\right),\quad \forall q \in \mcL(\mcK).
\end{equation}
The risk map $\widetilde{\mcR}$ implicitly reduces the average risk-aware control problem to an average risk evaluation on a cost generating Markov chain. We thus invoke a general Doeblin type condition for average risk evaluation on a Markov chain as follows, where we see $\mcK$ as the set of states of the Markov chain.

\begin{assumption}[Assumption 3.1, \cite{Shen2013sicon}]\label{assump-Poisson-Doeblin}
There exists a function $\tilde{w}:\mcK \to [0,+\infty)$, a monetary risk measure $\tilde{\nu}:\mcL(\mcK)\rightarrow \mbR$, and some constants $\tilde{K} > 0$, $\tilde{\gamma} \in (0,1)$, and $\tilde{\alpha} \in (0,1)$ such that:
    \begin{enumerate}[label=(\arabic*)]
        \item Let $\widetilde{\mcR}^{\#}_{x,a}(q) := \sup_{p\in\mcL(\mcX)}\{ \widetilde{\mcR}_{x,a}(q+p) - \widetilde{\mcR}_{x,a}(p)\}$ and $\overline{\widetilde{\mcR}^{\#}}_{x,a}(q) := \sup_{\lambda \neq 0} \frac{\widetilde{\mcR}^{\#}_{x,a}(\lambda q)}{\lambda}$. We have that
        \begin{equation*}
            \overline{\widetilde{\mcR}^{\#}}_{x,a} (\tilde{w}) \le \tilde{\gamma} \tilde{w}(x,a) + \tilde{K},\quad \forall (x,a) \in \mcK.
        \end{equation*}
        \item For all $q\ge p \in \mcL(\mcK)$, we have that:
        \begin{equation*}
            \inf_{(x,a)\in\mcK:\tilde{w}(x,a)\leq \tilde{R}} \{ \widetilde{\mcR}(q|x,a) - \tilde{\alpha} \tilde{\nu}(q) - \widetilde{\mcR}(p|x,a) + \tilde{\alpha} \tilde{\nu}(p)\} \ge 0,
        \end{equation*}
        for some $\tilde{R}>2\tilde{K}/(1-\tilde{\gamma})$.
    \end{enumerate}
\end{assumption}

\begin{lemma}\label{thm-tildeRproperty}
    If $\mcR$ satisfies Assumption \ref{assump-AROE-Doeblin}, then $\widetilde{\mcR}$ satisfies Assumption \ref{assump-Poisson-Doeblin}. 
\end{lemma}
\begin{proof}
    For the $\bar{\alpha}$ and $\nu$ satisfying Assumption \ref{assump-AROE-Doeblin}, define $\tilde{w}:=\bm{0}$, $\tilde{K}:=1$, $\tilde{\gamma}:=0.5$, $\tilde{\alpha}:= 0.5$, $\tilde{R}:=5>2\tilde{K}/(1-\tilde{\gamma})$, and monetary risk measure $\widetilde{\nu}(q):=\nu(\min_{a\in\mcA}q(\cdot,a))$.
    We have
    \[\overline{\widetilde{\mcR}^{\#}}_{x,a}(\bm{0})=\sup_{\lambda \neq 0} \frac{\widetilde{\mcR}^{\#}_{x,a}(\lambda \bm{0})}{\lambda}=\sup_{\lambda \neq 0} \frac{\widetilde{\mcR}^{\#}_{x,a}(\bm{0})}{\lambda} = \sup_{\lambda \neq 0} \frac{0}{\lambda}=0\leq 0.5\cdot0+1=\tilde{\gamma} \tilde{w}(x,a)+ \tilde{K},\]
    where we exploited the fact that:
    \[\widetilde{\mcR}^{\#}_{x,a}(\bm{0}) = \sup_{q\in\mcL(\mcK)}\{ \widetilde{\mcR}_{x,a}(q) - \widetilde{\mcR}_{x,a}(q)\}=0,\quad \forall(x,a)\in\mcK.\]
    
    Moreover, for all $q\ge p \in \mcL(\mcK)$, we have
        \begin{align*}
            &\min_{(x,a)\in\mcK, \tilde{w}(x,a)\leq \tilde{R}} \{ \widetilde{\mcR}(q|x,a) - \bar{\alpha} \tilde{\nu}(q) - \widetilde{\mcR}(p|x,a) + \bar{\alpha} \tilde{\nu}(p)\}\\
            &\; = \min_{(x,a)\in\mcK} \{ \widetilde{\mcR}(q|x,a) - \bar{\alpha} \tilde{\nu}(q) - \widetilde{\mcR}(p|x,a) + \bar{\alpha} \tilde{\nu}(p)\}\\
            &\; = \min_{(x,a)\in\mcK} \{ \mcR_{x,a}(\min_{a'} q(\cdot,a')) - \bar{\alpha} \nu(\min_{a'}q(\cdot,a')) - \mcR_{x,a}(\min_{a'}p(\cdot,a')) + \bar{\alpha} \nu(\min_{a'}p(\cdot,a'))\}\\
            &\;\geq 0,
        \end{align*}
    where the last inequality follows from Assumption \ref{assump-AROE-Doeblin} using $v(\cdot) := \min_{a'\in\mcA} q(\cdot,a')$ and $v'(\cdot) := \min_{a'\in\mcA} p(\cdot,a')$.
    The result follows.
\end{proof}

We now list the properties of span-seminorm contractive operators that we will use later on.
\begin{lemma}[Theorem 6.6.2, \cite{Puterman1994book}]\label{lemma-spanseminorm}
    Let $\mcT:\mcL(\Omega) \to \mcL(\Omega)$, for some finite space $\Omega$, be an operator that is span-seminorm contractive, i.e., there exists an $\bar{\alpha} \in [0,1)$ such that $\norm{\mcT(v) - \mcT(w)}_{sp} \le \bar{\alpha} \norm{v- w}_{sp}$, for all $v,w\in\mcL(\Omega)$. Then the followings are true:
    \begin{enumerate}[label=(\arabic*)]
        \item There exists a $v^* \in \mcL(\Omega)$ such that $\norm{\mcT (v^*) - v^*}_{sp} = 0$. Such $v^*$ is called the span-seminorm fixed point of the operator $\mcT$.
        \item For all $n \ge 0$, $\norm{\mcT^{n}(v) - v^*}_{sp} \le \bar{\alpha}^n \norm{v - v^*}_{sp}$.
        \item For any $v \in \mcL(\Omega)$, we have $\lim_{n\to\infty} \norm{\mcT^n(v) - v^*}_{sp} = 0$.
        \item Any two span-seminorm fixed points of $\mcT$ must differ by a constant.
    \end{enumerate}
\end{lemma}

In our proofs we will exploit the fact that $\widetilde{\mcR}$ is non-expansive and span-seminorm contractive.

\begin{lemma}\label{lemma-tildeRnonexpanSpancontract}
    If the risk map $\mcR$ satisfies Assumption \ref{assump-AROE-Doeblin},
    then both $\widetilde{\mcR}$ and $\mcH$ (from equation \eqref{algo-RQI-generalR}) are non-expansive under the infinity norm and span-seminorm contractive.
\end{lemma}
\begin{proof}
    From Proposition 3.6 in \cite{Shen2013sicon}, for any risk map $\widetilde{\mcR}$, which satisfies Assumption \ref{assump-Poisson-Doeblin} based on Lemma \ref{thm-tildeRproperty}, we have $|\widetilde{\mcR}_{x,a}(q + p) - \widetilde{\mcR}_{x,a}(p)| \le \overline{\widetilde{\mcR}^\#}_{x,a}(|q|)$, where $|q|(x):=|q(x)|$. Since $\overline{\mcR^\#}_{x,a}$ is a coherent risk measure (see Proposition 3.5 in \cite{Shen2013sicon}), we have for all $q,p\in\mcL(\mcK)$:
    \begin{align*}
        |\widetilde{\mcR}_{x,a}(q) - \widetilde{\mcR}_{x,a}(p)| \le \overline{\widetilde{\mcR}^\#}_{x,a}(|q-p|) \le \overline{\widetilde{\mcR}^\#}_{x,a}(\norm{q-p}_\infty e) =\norm{q-p}_\infty,
    \end{align*}
    where the second inequality follows from monotonicity of $\overline{\widetilde{\mcR}^\#}_{x,a}$, while the last equality comes from translation invariance and normalization of $\overline{\widetilde{\mcR}^\#}_{x,a}$.

    The span-seminorm contraction straightforwardly follows from Theorem 3.11 in \cite{Shen2013sicon} given the fact that $\widetilde{\mcR}$ satisfies Assumption \ref{assump-Poisson-Doeblin} as established in Lemma \ref{thm-tildeRproperty}. 

    These properties carry directly to $\mcH$ since
    \[ \norm{\mcH (q) - \mcH (p)}_\infty = \norm{\widetilde{\mcR}(q|\cdot)-\widetilde{\mcR}(p|\cdot)}_\infty \le \norm{q-p}_\infty, \]
    and
    \[ \norm{\mcH (q) - \mcH (p)}_{sp} = \norm{\widetilde{\mcR}(q|\cdot)-\widetilde{\mcR}(p|\cdot)}_{sp}\leq \bar{\alpha}\norm{q-p}_{sp} \]
    for some $\bar{\alpha}\in[0,1)$. This completes the proof.
\end{proof}

We are now ready prove the convergence of the risk-aware RQI algorithm \eqref{algo-RQI-generalR}.
\begin{proof}[Proof of Theorem \ref{thm-RQI-generalR}]
    Define $\bar{V}_n(x) := \min_{a\in\mcA} Q_n(x,a)$, $\forall x\in \mcX$. 
    Taking minimum over $a$ on both sides of \eqref{algo-RQI-generalR}, we obtain
    \begin{align*}
        \bar{V}_{n+1}(x) = \min_{a\in\mcA}\{ c(x,a) + \mcR_{x,a}(\bar{V}_n)\} -f(Q_n) = \mcG(\bar{V}_n)(x) - f(Q_n),\;\forall x\in\mcX,
    \end{align*}
    where $\mcG$ is the risk-aware Bellman optimality operator.
    If $Q_n$ converges to some fixed point $Q_\infty$ of \eqref{algo-RQI-generalR} under the infinity norm, we have
    \begin{equation}\label{eq-PRQI-1}
        \min_{a\in\mcA} Q_\infty(x,a) = \min_{a\in\mcA}\left\{ c(x,a) + \mcR_{x,a}\left(\min_{a'\in\mcA} Q_\infty(\cdot,a')\right)\right\} - f(Q_\infty),\quad \forall x\in\mcX.
    \end{equation}
    Notice that $\min_{a\in\mcA} Q_\infty(\cdot,a) \in \mcL(\mcX)$, $f(Q_\infty) \in \mbR$ and $Q_\infty$ satisfies \ref{eq-PRQI-1}, we conclude that $(\min_{a\in\mcA} Q_\infty(\cdot,a),f(Q_\infty))$ is a pair of solution to the AROE \eqref{eq-AROE-generalR}. By Theorem \ref{thm-AROE-generalR}, $f(Q_\infty) = g^*$. Therefore, 
    We are left with the task to show that $Q_n$ converges to some unique fixed point $Q_\infty$ of \eqref{algo-RQI-generalR}.

    To analyze the convergence, consider the augmented risk map defined in \eqref{eq-augmentedriskmap}. Then algorithm \eqref{algo-RQI-generalR} can be equivalently written as:
    \begin{equation*}
        Q_{n+1}(x,a) = c(x,a) + \widetilde{\mcR}_{x,a}\left( Q_n\right) - f(Q_n),\quad \forall (x,a) \in \mcK.
    \end{equation*}
    given the translation invariance of $\mcR_{x,a}$. Its convergence can be associated to the convergence of an average risk estimator on a Markov chain, with $\mcK$ as the state space, under the risk map $\widetilde{\mcR}$ on the $\mcK$ outcome space, which is studied in Section 3 of \cite{Shen2013sicon}. Indeed, given that $\widetilde{\mcR}$ satisfies Assumption \ref{assump-Poisson-Doeblin}, Theorem 3.14 (i) in \cite{Shen2013sicon} already establishes that the Poisson equation, 
    \begin{equation}\label{eq-Poisson}
        c(x,a) + \widetilde{\mcR}_{x,a}(q) = g + q(x,a),\quad \forall(x,a)\in\mcK,
    \end{equation}
    has a solution $(q^*,\tilde{g}^*)$, where $\tilde{g}^*$ is unique.
    
    By Lemma \ref{lemma-tildeRnonexpanSpancontract}, the risk-aware Bellman optimality operator $\mcH$ (see \eqref{algo-RQI-generalR}) is span-seminorm contractive. This implies, based on  Lemma \ref{lemma-spanseminorm}, that $\mcH$ has a span-seminorm fixed point, i.e, there exists $q^* \in \mcL(\mcK)$ and $\tilde{g}^*\in\mbR$ such that $\norm{\mcH(q^*) - q^*}_{sp} = 0$ and $q^*+\tilde{g}^* = \mcH(q^*)$. The latter implies that $(q^*,\tilde{g}^*)$ satisfies the Poisson equation \eqref{eq-Poisson} and that $\lim_{n\rightarrow \infty} \norm{\mcH^n(q)-q^*}_{sp} = 0$, for any $q\in\mcL(\mcK)$, due to the span-seminorm contraction property of $\mcH$ (Lemma \ref{lemma-spanseminorm}).

    One can further show that $\mcH^{n+1}(q)-\mcH^{n}(q)\rightarrow \tilde{g}^*$, for any $q\in\mcL(\mcK)$, using
    \begin{align*}
        &\norm{\mcH^{n+1}(q)-\mcH^{n}(q)-\tilde{g}^*}_\infty \\
        &= \inf_g \norm{\mcH(q^*+g+\mcH^n(q)-g-q^*) - (q^*+g+\mcH^n(q)-g-q^*)-\tilde{g}^*}_\infty\\
        &=\inf_g \norm{\mcH(q^*+\mcH^n(q)-g-q^*) + g - q^* - g -\mcH^n(q)+g+q^*-\tilde{g}^*}_\infty\\
        &\leq \inf_g \left\{ \norm{\mcH(q^*+\mcH^n(q)-g-q^*) -\mcH(q^*)}_\infty + \norm{\mcH(q^*)-q^*-\tilde{g}^*}_\infty + \norm{\mcH^n(q)-g-q^*}_\infty\right\}\\
        &\leq \inf_g \left\{\norm{\mcH^n(q)-g-q^*}_\infty + \norm{q^*+\tilde{g}^*-q^*-\tilde{g}^*}_\infty + \norm{\mcH^n(q)-g-q^*}_\infty\right\}\\
        &= 2\inf_g \norm{\mcH^n(q)-g-q^*}_\infty =\norm{\mcH^n(q)-q^*}_{sp}.
    \end{align*}
    where the second equality comes from translation invariance, the first inequality comes from the triangular inequality, the second inequality follows from $\mcH$ being non-expansive (see Lemma \ref{lemma-tildeRnonexpanSpancontract}), and finally the last equality is proved as Lemma 3.9 in \cite{Shen2013sicon}.
    Hence, we must have that
    $\lim_{n\rightarrow 0} \norm{\mcH^{n+1}(q)-\mcH^{n}(q)-\tilde{g}^*}_\infty \leq \lim_{n\rightarrow \infty} \norm{\mcH^n(q)-q^*}_{sp} = 0$.

    We now wish to analyze the convergence of the process $\{Q_{n}\}_{n=0}^\infty$ produced by our algorithm. To do so, consider the process $U_{n+1} := \mcH(U_n)$ with $U_0 := Q_0$, for which we know that $\norm{\tilde{g}_n - \tilde{g}^*}_\infty \to 0$ as $n\to\infty$, where $\tilde{g}_n := U_{n+1} - U_{n}$, for $n\ge 0$. One can actually establish by induction that $Q_{n}=U_n-f(U_{n-1})$ for all $n\geq 1$. Namely, start at $n=1$ where 
    \[Q_1=\mcH (Q_0) - f(Q_0)=\mcH (U_0) - f(U_0)= U_1- f(U_0).\]
    Then iteratively assuming that $Q_n=U_n-f(U_{n-1})$, one can confirm that:
    \begin{align*}
        Q_{n+1}&=\mcH(Q_n)-f(Q_n)=\mcH (U_n-f(U_{n-1})) - f(U_n-f(U_{n-1}))\\
        &=\mcH (U_n)  -f(U_{n-1})-f(U_n)+f(U_{n-1})=\mcH (U_n) - f(U_n)=U_{n+1}-f(U_n).        
    \end{align*}
    This relation can be used to establish that
    \begin{align*}
        \norm{Q_{n+1} - Q_n}_\infty &= \norm{U_{n+1} - f(U_{n}) - U_n + f(U_{n-1})}_\infty\\
        &= \norm{\tilde{g}_n - f(U_{n}) + f(U_{n-1} + \tilde{g}^*) - f(U_{n-1} + \tilde{g}^*) + f(U_{n-1})}_\infty\\
        &= \norm{\tilde{g}_n - \tilde{g}^* - f(U_{n}) + f(U_{n-1} + \tilde{g}^*)}_\infty\\
        &\le \norm{\tilde{g}_n - \tilde{g}^*}_\infty + \norm{f(U_{n}) - f(U_{n-1} + \tilde{g}^*)}_\infty\\
        &\le \norm{\tilde{g}_n - \tilde{g}^*}_\infty + \tilde{L}\norm{U_{n} - U_{n-1} - \tilde{g}^*}_\infty\\
        & = \norm{\tilde{g}_n - \tilde{g}^*}_\infty + \tilde{L}\norm{\tilde{g}_{n-1} - \tilde{g}^*}_\infty.        
    \end{align*}    
    where the third equality follows from the translation invariance property of $f$ imposed in Assumption \ref{assump-RQI-f} (i), while the final inequality arises from the Lipschitz property of $f$ in in Assumption \ref{assump-RQI-f} (ii), with $\tilde{L} \ge 0$ as the Lipschitz constant. When $n\to\infty$, we have shown that $\norm{\tilde{g}_n - \tilde{g}^*}_\infty = \norm{\mcH^{n+1} (Q_0) - \mcH^n (Q_0) - \tilde{g}^*}_\infty \to 0$. Therefore we can conclude that $Q_n$ converges to some $Q_\infty$ and $f(Q_n)$ converges to $\tilde{g}^*$ as $n\to\infty$, i.e., $(Q_\infty,f(Q_\infty))$ satisfies the Poisson equation \eqref{eq-Poisson}.

    Finally, we show that such $Q_\infty$ is independent of $Q_0$ for a fixed $f$. Since $Q_n \to Q_\infty$ as $n\to\infty$, from \eqref{algo-RQI-generalR}, we obtain that
    \begin{equation}\label{algo-RQI-fixedpoint}
        Q_\infty(x,a) = \mcH(Q_\infty)(x,a) - f(Q_\infty),\quad \forall (x,a) \in \mcK.
    \end{equation}
    Notice that $f(Q_\infty) = \tilde{g}^*$ is a constant, this implies that $Q_\infty$ is a span-seminorm fixed point of $\mcH$. Suppose $\widetilde{Q}_\infty$ is another solution to \eqref{algo-RQI-fixedpoint}. Then by Lemma \ref{lemma-spanseminorm}(iv),  $Q_\infty$ and $\widetilde{Q}_\infty$ only differs by a constant.  
    Yet, we know that $\tilde{g}^*=f(\widetilde{Q}_\infty) = f(Q_\infty + r) = f(Q_\infty) + r = \tilde{g}^* + r$, which implies that $r=0$ and that $Q_\infty=\widetilde{Q}_\infty$. We therefore conclude that $Q_\infty$ is unique. 

    From the analysis in the first part, we conclude that $\tilde{h}^*:=\min_{a\in\mcA} Q_\infty(\cdot,a)$ and $\tilde{g}^*:=f(Q_\infty)$ identify a  solution pair to the AROE \eqref{eq-AROE-generalR} and thus $f(Q_\infty)=g^*$.
\end{proof}

We now turn to establishing Theorem \ref{thm-RVI-generalR}.
\begin{proof}[Proof of Theorem \ref{thm-RVI-generalR}]
    Let $\tilde{f}:\mcL(\mcX) \to \mbR$ be a function satisfying Assumption \ref{assump-RVI-f}. Define a function $\hat{f}:\mcL(\mcK) \to \mbR$ as $\hat{f}(Q_n) := \tilde{f}(\min_{a\in\mcA} Q_n(\cdot,a))$. It is easy to verify that $\hat{f}$ satisfies Assumption \ref{assump-RQI-f}. 

    Consider the risk-aware RQI \eqref{algo-RQI-generalR} with function $\hat{f}$ starting with $Q_0$. Define $\bar{V}_n(x) := \min_{a\in\mcA} Q_n(x,a)$, for all $x\in \mcX$. 
    Taking minimum over $a$ on both sides of \eqref{algo-RQI-generalR}, we obtain
    \begin{align*}
        \bar{V}_{n+1}(x) &= \min_{a\in\mcA}\{ c(x,a) + \mcR_{x,a}(\bar{V}_n)\} -\hat{f}(Q_n) = \mcG(\bar{V}_n)(x) - \hat{f}(Q_n)\\
        &=\mcG(\bar{V}_n)(x) - \tilde{f}(\bar{V}_n),\quad \forall x \in \mcX,
    \end{align*}
    where $\mcG$ is the risk-aware Bellman optimality operator defined in \eqref{algo-RVI-generalR}. This is exactly the risk-aware RVI algorithm with initial value function $V_0(x) = \min_{a\in\mcA}Q_0(x,a)$, for all $x\in\mcX$.
    
    By Theorem \ref{thm-RQI-generalR}, the RQI algorithm converges, i.e., $Q_n$ converges to some unique fixed point of \eqref{algo-RQI-generalR} called $Q^*$. Hence, $\bar{V}_n$ converges to some $\bar{V}^* := \min_{a\in\mcA}Q^*(\cdot,a)$.
    Since we can always design a $Q_0$ such that $\min_{a\in\mcA}Q_0(x,a) = \bar{V}_0(x)$ for any $\bar{V}_0 \in \mcL(\mcX)$. 
    We conclude that for any initial value $V_0$, the risk-aware RVI algorithm \eqref{algo-RVI-generalR} converges to some $V^*$.

    Using the same reasoning as in Lemma \ref{lemma-tildeRnonexpanSpancontract}, we can conclude that the risk-aware Bellman optimality operator $\mcG$ is non-expansive under the infinity norm and contractive with respect to the span-seminorm. Based on Assumption \ref{assump-RVI-f} and the preceding derivation, taking the limit on both sides of \eqref{algo-RVI-generalR} yields the equation \(V^*(x) = \mcG(V^*)(x) - \tilde{f}(V^*)\) for all \(x \in \mcX\). This implies that \((V^*, \tilde{f}(V^*))\) identifies a pair of solution to the AROE \eqref{eq-AROE-generalR}, leading to \(\tilde{f}(V^*) = g^*\) and \(V^*\) serves as a fixed point of \eqref{algo-RVI-generalR}. The uniqueness of \(V^*\) follows from the same argument used to establish the uniqueness of \(Q^*\) in the proof of Theorem \ref{thm-RQI-generalR}. This completes the proof.
\end{proof}

\subsection{Proof of Theorem \ref{thm-RVIQ-generalR}}
In this section, we use the ODE analysis of stochastic approximation to prove the convergence of the risk-aware RVI Q-learning algorithm. 

Define an operator $H:\mcL(\mcK) \to \mcL(\mcK)$ as
\[ H(q)(x,a) := \mcR_{x,a}\left(c(x,a) + \min_{a'\in\mcA} q(\cdot,a')\right) - f(q) - q(x,a),\quad \forall q\in\mcL(\mcK). \]
Then the update of the risk-aware RVI Q-learning can be written as
\begin{align*}
    Q_{n+1} &= Q_n + \gamma(n)(\hat{\mcH}(Q_n) - f(Q_n) - Q_n)\\
    &=Q_n + \gamma(n)(H(Q_n)+ \hat{\mcH}(Q_n) - H(Q_n) - f(Q_n) - Q_n).
\end{align*}
Hence we have the stochastic approximation iteration:
\begin{equation}\label{eq-RVIQ-SA}
    Q_{n+1} = Q_n + \gamma(n)(H(Q_n) + M_{n+1}),
\end{equation}
where $M_{n+1} := \hat{\mcH}(Q_n) - \mcH(Q_n)$ is the noise term. 

The classical approach to analyzing stochastic approximation using ODEs involves examining the stability of the equilibrium of a corresponding ODE related to \eqref{eq-RVIQ-SA}:
\begin{equation}\label{eq-ODE-p}
    \dot{p}_t = H(p_t).
\end{equation}
If the ODE \eqref{eq-ODE-p} has a unique globally asymptotically stable equilibrium point $p^*$, then under certain conditions, the stochastic approximation \eqref{eq-RVIQ-SA} converges, with $Q_n \to Q^* = p^*$ almost surely (see Theorem 2.2 of \cite{Borkar2000sicon}). Notice that if such $p^*$ exists, then $Q^*$ is a solution to the AROE \eqref{eq-RQI-AROE}, which implies that $(\min_{a'\in\mcA} p^*(\cdot,a'),f(p^*))$ is a pair of solution to the AROE \eqref{eq-AROE-generalR}. Then the result of Theorem \ref{thm-RVIQ-generalR} follows easily.

Analyzing the stability of the equilibrium point of ODE \eqref{eq-ODE-p} can sometimes be challenging. A common approach is to employ a time-averaging technique to smooth out perturbations and examine the stability of the origin in the limiting ODE.
Namely, we define an operator $H_s:\mcL(\mcK) \to \mcL(\mcK)$ as $H_s(Q) := \frac{1}{s}H(sQ)$, with $s \ge 1$ and consider the ODE:
\begin{equation}\label{eq-ODE-phi}
    \dot{\phi_t} = H_s(\phi_t).
\end{equation}

Following \cite{Borkar2000sicon}, to establish the convergence of the stochastic approximation \eqref{eq-RVIQ-SA}, we outline the sufficient conditions that are needed to be verified:
\begin{enumerate}[label=(\roman*)]
    \item\label{cond:i:Lipschitz} The function $H$ is Lipschitz.
    \item\label{cond:iii:martingale}  The sequence $\{M_n, \mcF_n : n\ge 1\}$ with $\mcF_n := \sigma(Q_i,M_i,i\le n)$ is a martingale difference sequence. Moreover, there exists some $C_0 < \infty$ and for any initial condition $Q_0 \in \mcL(\mcK)$ we have almost surely,
    \[ \mbE[\norm{M_{n+1}}_\infty^2 |\mcF_n] \le C_0(1 + \norm{Q_n}_\infty^2),\quad n\ge 0. \]
    \item\label{cond:iv:stepsize}  The step size satisfies the Robbins-Monro condition (see Assumption \ref{assump-RobbinsMonro}).
    \item\label{cond:v:bounded}  For any initial condition $Q_0 \in \mcL(\mcK)$, the iteration is bounded almost surely, i.e., $\sup_n \norm{Q_n}_\infty < \infty$, almost surely.
    \item\label{cond:vi:stable}  The ODE \eqref{eq-ODE-p} has a unique globally asymptotically stable equilibrium point.
    \item\label{cond:ii:Hinf}  The limit $H_\infty(Q) := \lim_{s\to\infty} H_s(Q)$ exists and the convergence is uniform on compact sets, and the ODE 
    \begin{equation}\label{eq-ODE-phi_infinity}
        \dot{\phi}_t = H_\infty(\phi_t),
    \end{equation}
    has the origin as an asymptotically stable equilibrium.
\end{enumerate}

Following Theorem 2.2 of \cite{Borkar2000sicon}, if conditions \ref{cond:i:Lipschitz}, \ref{cond:iii:martingale}, \ref{cond:iv:stepsize}, \ref{cond:v:bounded} and \ref{cond:vi:stable} hold, then the stochastic approximation \eqref{eq-RVIQ-SA} converges almost surely to the unique globally stable equilibrium point of the ODE \eqref{eq-ODE-p}, which is a solution to the AROE \eqref{eq-RQI-AROE}, thus confirming our theorem.

The Lipschitz property \ref{cond:i:Lipschitz} is straightforward to verify, as the risk measure \(\mcR_{x,a}\) and the function \(f\) are both Lipschitz (see Lemma \ref{lemma-tildeRnonexpanSpancontract} and Assumption \ref{assump-RQI-f}). As stated in Theorem 2.1 of \cite{Borkar2000sicon}, the almost sure boundedness condition \ref{cond:v:bounded} follows from conditions \ref{cond:i:Lipschitz}, \ref{cond:ii:Hinf}, \ref{cond:iii:martingale} and \ref{cond:iv:stepsize}, where condition \ref{cond:iv:stepsize} is automatically satisfied by Assumption \ref{assump-RobbinsMonro}.  
Thus, the remainder of this section focuses on verifying conditions \ref{cond:i:Lipschitz}, \ref{cond:iii:martingale}, \ref{cond:vi:stable} and \ref{cond:ii:Hinf}.

\subsubsection{Condition \ref{cond:i:Lipschitz}}
\begin{lemma}\label{lemma-HLipschitz}
    $H$, $H_s$ and $H_\infty$, if it exists, are Lipschitz and have the same Lipschitz constant.
\end{lemma}
\begin{proof}
    Following Lemma \ref{lemma-tildeRnonexpanSpancontract}, $\widetilde{\mcR}$ is non-expansive. Hence, for any $Q_1, Q_2 \in \mcL(\mcK)$, with Assumption \ref{assump-RQI-f}, we have
    \begin{align*}
        H(Q_1)(x,a) - H(Q_2)(x,a) 
        &= \widetilde{\mcR}_{x,a}(Q_1) - \widetilde{\mcR}_{x,a}(Q_2) - f(Q_1) + f(Q_2) - Q_1(x,a)-Q_2(x,a)\\ 
        &\le \norm{Q_1 - Q_2}_\infty + \tilde{L}\norm{Q_1 - Q_2}_\infty + \norm{Q_1-Q_2}_\infty\\
        &= (2 + \tilde{L})\norm{Q_1 - Q_2}_\infty,
    \end{align*}
    where $\tilde{L}$ is the Lipschitz constant for $f$.
    Similarly, we obtain $H(Q_2)(x,a) - H(Q_1)(x,a) \le (2+\tilde{L})\norm{Q_2 - Q_1}_\infty$. Hence $H$ is Lipschitz with Lipschitz constant $2 + \tilde{L}$.

    Meanwhile,
    \begin{align*}
        &H_s(Q_1)(x,a) - H_s(Q_2)(x,a) \\
        &= \frac{1}{s} \Big(\mcR_{x,a}\big(c(x,a) + \min_{a'\in\mcA} s Q_1(x,a)\big) - \mcR_{x,a}\big(c(x,a) + \min_{a'\in\mcA} s Q_2(x,a)\big) \\
        &\qquad - f(sQ_1) + f(sQ_2) - sQ_1(x,a) + sQ_2(x,a)\Big)\\
        &\le \frac{1}{s} (s\norm{Q_1 - Q_2}_\infty + s \norm{Q_1 - Q_2}_\infty + s\norm{Q_1-Q_2}_\infty)\\
        & = (2 + \tilde{L})\norm{Q_1 - Q_2}_\infty.
    \end{align*}
    Similarly, we obtain $H_s(Q_2)(x,a) - H_s(Q_1)(x,a) \le (2+\tilde{L})\norm{Q_2 - Q_1}_\infty$. Hence $H_s$ is Lipschitz with Lipschitz constant $2 + \tilde{L}$.
    Similarly, if $H_\infty(Q) := \lim_{s\to\infty} H_s(Q)$ exists, $H_\infty$ is also Lipschitz with Lipschitz constant $2+\tilde{L}$.
\end{proof}

\subsubsection{Condition \ref{cond:iii:martingale}}
We check that $M_n$ is a martingale difference sequence that satisfies \ref{cond:iii:martingale}.

\begin{lemma}\label{lemma-martingale}
    Under Assumption \ref{assump-unbiased}, for all $n=0,1,...$, we have $\mbE[M_{n+1} |\mcF_n]=0$ almost surely and there exists some $C_0 < \infty$ such that for any initial condition $Q_0 \in \mcL(\mcK)$ we have
    \[ \mbE[\norm{M_{n+1}}_\infty^2 |\mcF_n] \le C_0(1 + \norm{Q_n}_\infty^2),\quad n\ge 0,\quad \mbox{a.s.} \]
\end{lemma}
\begin{proof}
By Assumption \ref{assump-unbiased}, it is easy to see that
\[ \mbE[\hat{\mcH}(Q_n)|\mcF_n] = \mcH(Q_n),\; \mbox{a.s}, \quad \Var[\hat{\mcH}(Q_n)(x,a)|\mcF_n] \le C(1 + \norm{Q_n}_\infty^2), \; \forall (x,a) \in \mcK, \mbox{a.s.}, \]
for some constant $C > 0$. 
Then by definition,
\begin{align*}
    \mbE[M_{n+1}|\mcF_n] = \mbE[\hat{\mcH}(Q_n) - \mcH(Q_n)|\mcF_n] = \mbE[\hat{\mcH}(Q_n)|\mcF_n] - \mcH(Q_n) = 0,\quad \mbox{a.s.}
\end{align*}
Meanwhile, for any $(x,a) \in \mcK$, from the definition of variance, we have for all $(x,a)\in\mcK$ that almost surely
\begin{align*}
    \mbE[(M_{n+1}(x,a))^2|\mcF_n] &= \mbE[(\hat{\mcH}(Q_n)(x,a) - \mcH(Q_n)(x,a))^2|\mcF_n]\\
&= \mbE[(\hat{\mcH}(Q_n)(x,a) - \mbE[\hat{\mcH}(Q_n)(x,a)|\mcF_n])^2|\mcF_n]\\    
    &= \Var[\hat{\mcH}(Q_n)(x,a)|\mcF_n] \\
    &\le C(1+\norm{Q_n}_\infty^2) .
\end{align*}
This implies that
\[ \mbE[\norm{M_{n+1}}_2^2 |\mcF_n] \le |\mcK| C(1+\norm{Q_n}_\infty^2),\quad \mbox{a.s.} \]
From the $L_p$-norm inequality, we have
\begin{align*}
    \mbE[\norm{M_{n+1}}_\infty^2|\mcF_n] \le \mbE[\norm{M_{n+1}}_2^2 |\mcF_n] \le  |\mcK| C(1+\norm{Q_n}_\infty^2) =: C_0 (1+\norm{Q_n}_\infty^2),\quad \mbox{a.s.},
\end{align*}
for some constant $C_0 < \infty$. This completes the proof.
\end{proof}

\subsubsection{Condition \ref{cond:vi:stable}}
In this subsection, we always assume Assumption \ref{assump-RQI-f} holds for all the lemmas.
To prove \ref{cond:vi:stable}, we need to analyze the stability of the equilibrium of ODE \eqref{eq-ODE-p}, which is quite difficult as there exists nonlinear terms $\mcR_{x,a}$ and $f$. Following \cite{Abounadi2001sicon}, we also analyze the behavior of an ODE where we replace $f(Q)$ with a constant $g^*$:
\begin{equation}\label{eq-ODE-q}
    \dot{q}_t = \tilde{H}(q_t),
\end{equation}
where $\tilde{H}(q) := \mcH(q) - g^* - q$, $\forall q \in \mcL(\mcK)$. Clearly, the fixed point $q^*$ of ODE \eqref{eq-ODE-q}, together with $g^*$, is a solution to the AROE \eqref{eq-RQI-AROE}. Hence, under Assumption \ref{assump-AROE-Doeblin}, the set of fixed point of \eqref{eq-ODE-q} is not empty and by lemmas \ref{lemma-spanseminorm} and \ref{lemma-tildeRnonexpanSpancontract}, the fixed points differ by a constant. We conclude on the characteristic of the set of the equilibrium points of ODE \eqref{eq-ODE-q} using the following lemma.
\begin{lemma}\label{lemma-Gset}
    The set $G$ of equilibrium of ODE \eqref{eq-ODE-q} satisfies $G = \{ q: q = \bar{q}^* + r, r\in\mbR\}$, where $\bar{q}^*$ is the only solution to the AROE \eqref{eq-RQI-AROE} that satisfies $f(\bar{q}^*) = g^*$.
\end{lemma}
\begin{proof}
    It is evident that any solution $q^*$ to the AROE \eqref{eq-RQI-AROE} satisfies $0 = \tilde{H}(q^*)$, indicating that $q^*$ is an equilibrium point for the ODE \eqref{eq-ODE-q}. According to Theorem \ref{thm-AROE-generalR}, under Assumption \ref{assump-AROE-Doeblin}, the set of equilibrium points is non-empty. Similarly, for any equilibrium point $\tilde{q}\in G$, we have $0 = \tilde{H}(\tilde{q})$, which satisfies the AROE \eqref{eq-RQI-AROE}, implying that $\tilde{q}$ is a solution to the AROE \eqref{eq-RQI-AROE}, i.e., a span-seminorm fixed point of $\mcH$. By Lemma \ref{lemma-spanseminorm}, each fixed point differs only by a constant. 
    Therefore, we conclude that $G = \{q: q = \tilde{q}+ r, r \in \mbR\}$ for some equilibrium point $\tilde{q}$.
    
    Now suppose $f(\tilde{q}) = m$ for some constant $m$. Then by Assumption \ref{assump-RQI-f}, $f(\tilde{q} + g^* - m) = f(\tilde{q}) + g^*-m = g^*$. Hence, we can always find a $\bar{q}^* := \tilde{q}+ g^* - m$ satisfying $f(\bar{q}^*) = g^*$. By definition, $\bar{q}^* \in G$, therefore is a solution to the AROE \eqref{eq-RQI-AROE}. 
\end{proof}

For notation simplicity, define
\[ \bar{\mcH}(q) := \mcH (q) - f(q),\quad \tilde{\mcH}(q) := \mcH(q) - g^*,\quad q \in \mcL(\mcK). \]
Then for the two ODEs \eqref{eq-ODE-p} and \eqref{eq-ODE-q}, we have
\[ \dot{p}_t = H(p_t) = \bar{\mcH}(p_t) - p_t,\quad \dot{q}_t = \tilde{H}(q_t) = \tilde{\mcH}(q_t) - q_t. \]

Since $\mcH$ is non-expansive (see Lemma \ref{lemma-tildeRnonexpanSpancontract}), $\tilde{\mcH}$ is also non-expansive. From Theorem 3.1 of \cite{Borkar1997} (also see Lemma 3.1 of \cite{Abounadi2001sicon}), the ODE \eqref{eq-ODE-q} has a unique trajectory that may depend on the initial point $q_0$ and converges to some equilibrium point $q^*$. We conclude as the following lemma.

\begin{lemma}\label{lemma-Abounadi}
    Let $q_t$ be a solution of ODE \eqref{eq-ODE-q}. Then $q_t \to q^*$ as $t\to\infty$ for some equilibrium point $q^*$ of \eqref{eq-ODE-q} that may depend on $Q_0$. 
    Moreover, $q^*=\bar{q}^* + \bar{r}$ for some $\bar{r}\in\mbR$, where $\bar{q}^*$ is defined in Lemma \ref{lemma-Gset}.
\end{lemma}
\begin{proof}
    The convergence result follows from \cite{Abounadi2001sicon} Lemma 3.1. Then the result follows by applying Lemma \ref{lemma-Gset}.
\end{proof}

Following the property of $f$, we can show that the equilibrium point of ODE \eqref{eq-ODE-p} is unique and is also included in the set of equilibrium points of ODE \eqref{eq-ODE-q}.
\begin{lemma}\label{lemma-ODE-unique}
    The point $\bar{q}^*$ is the unique equilibrium point of ODE \eqref{eq-ODE-p}.
\end{lemma}
\begin{proof}
    Based on Lemma \ref{lemma-Gset}, since $f(\bar{q}^*) = g^*$, we have $\bar{\mcH}(\bar{q}^*) = \tilde{\mcH}(\bar{q}^*) = \bar{q}^*$, which means that $\bar{q}^*$ is an equilibrium point for \eqref{eq-ODE-p}. Conversely, if there exists some $\tilde{p}$ such that $\bar{\mcH}\tilde{p} = \tilde{p}$, by definition, the solution of the above equation satisfies the AROE \eqref{eq-RQI-AROE}. By Theorem \ref{thm-RQI-generalR}, $f(\tilde{p}) = g^*$.
    Therefore, we have $\tilde{p} = \bar{\mcH}(\tilde{p}) = \tilde{\mcH}(\tilde{p})$, which means $\tilde{p}$ is also an equilibrium for \eqref{eq-ODE-q}. By Lemma \ref{lemma-Gset}, $\tilde{p} = \bar{q}^* + \bar{r}$ for some $\bar{r}\in\mbR$. Then we have $g^* = f(\tilde{p}) = f(\bar{q}^* + r) = g^* + r$. This implies that $r = 0$.
    Therefore, $\bar{q}^*$ is a unique equilibrium point for ODE \eqref{eq-ODE-p}.
\end{proof}

The next result shows that the trajectory of ODE \eqref{eq-ODE-p} and ODE \eqref{eq-ODE-q} differs only by a constant function.
\begin{lemma}\label{lemma-ODE-r}
    Let $p_t$ and $q_t$ be the solutions to the ODEs \eqref{eq-ODE-p} and \eqref{eq-ODE-q},
    with the same initial value $p_0(x,a) = q_0(x,a) = Q_0(x,a)$. Then we have
    \[ p_t(x,a) = q_t(x,a) + r_t, \quad \forall (x,a) \in \mcK, \]
    where $r_t$ is a scalar function satisfying
    \[ \dot{r}_t = -r_t + g^* - f(q_t). \]
\end{lemma}
\begin{proof}
    Notice that $\bar{\mcH}(Q) = \tilde{\mcH}(Q) + (g^* - f(Q))$. Then from the variation of constants formula, we have that
    \begin{align*}
        p_t(x,a) &= q_0(x,a) \mre^{-t} + \int_0^t \mre^{-(t-s)} \tilde{\mcH}(p_s(x,a))ds + \int_0^t \mre^{-(t-s)}(g^* - f(p_s) ds,\\
        q_t(x,a) &= q_0(x,a) \mre^{-t} + \int_0^t \mre^{-(t-s)} \tilde{\mcH}(q_s(x,a))ds.
    \end{align*}
    The maximal and minimal components of $p_t - q_t$ can be bounded by
    \begin{align*}
        \max_{(x,a)\in\mcK} \{ p_t(x,a) - q_t(x,a)\} &\le \int_0^t \mre^{-(t-s)} \max_{(x,a)\in\mcK} \{ \tilde{\mcH}(p_s)(x,a) - \tilde{\mcH}(q_s)(x,a)\} ds \\
        &\qquad + \int_0^t \mre^{-(t-s)}(g^* - f(p_s)) ds,\\
        \min_{(x,a)\in\mcK} \{ p_t(x,a) - q_t(x,a)\} &\ge \int_0^t \mre^{-(t-s)} \min_{(x,a)\in\mcK} \{ \tilde{\mcH}(p_s)(x,a) - \tilde{\mcH}(q_s)(x,a)\} ds \\
        &\qquad + \int_0^t \mre^{-(t-s)}(g^* - f(p_s)) ds.
    \end{align*}
    Hence, we have
    \begin{align*}
        \norm{p_t - q_t}_{sp} &\le \int_0^t \mre^{-(t-s)} \norm{\tilde{\mcH}(p_s) - \tilde{\mcH}(q_s)}_{sp} ds\\
        &\le \int_0^t \mre^{-(t-s)} \norm{p_s - q_s}_{sp} ds.
    \end{align*}
    The inequality is from the fact that $\mcH$ is span-seminorm contractive (see Lemma \ref{lemma-tildeRnonexpanSpancontract}).
    By the Gronwall inequality, we have $\norm{p_t - q_t}_{sp} = 0$. This implies that there exists some scalar function $r_t$ such that $p_t(x,a) = q_t(x,a) + r(t)$ for all $(x,a) \in \mcK$, with $r(0) = 0$.

    Since $\tilde{\mcH}(p_t) = \tilde{\mcH}(q_t + r_t) = \tilde{\mcH}(q_t) + r_t$ and $f(p_t) = f(q_t + r_t) = f(q_t) + r_t$.
    Then the differential of $r_t$ is
    \begin{align*}
        \dot{r}_t e = \dot{p}_t - \dot{q}_t 
        =\tilde{\mcH}(p_t) + g^* - f(p_t) - p_t - \tilde{\mcH}(q_t) + q_t
        = (-r_t + g^* - f(q_t))e.
    \end{align*}
    This completes the proof.
\end{proof}

The following lemma shows that $\bar{q}^*$ is the unique globally asymptotically stable equilibrium point of ODE \eqref{eq-ODE-p}.

\begin{lemma}\label{lemma-ODE-GASE}
    $\bar{q}^*$ is the unique globally asymptotically stable equilibrium point of ODE \eqref{eq-ODE-p}.
\end{lemma}
\begin{proof}
    From Lemma \ref{lemma-ODE-r}, by the variation of constant formula, we have $r_t = \int_0^t \mre^{-(t-s)}(g^* - f(q_t)) ds$. By Lemma \ref{lemma-Abounadi}, we have $q_t \to q^* \in G$. Then we have $r_t \to g^* - f(q^*)$ so that $p_t \to q^* + (g^* - f(q^*))$, which must coincide with $\bar{q}^*$ since by Lemma \ref{lemma-ODE-unique}, it is the only equilibrium point of ODE \eqref{eq-ODE-p}. Next we show the Lyapunov stability of $\bar{q}^*$. Notice that
    \begin{align*}
        \norm{p_t - \bar{q}^*}_\infty &\le \norm{q_t - \bar{q}^*}_\infty + |r_t|\\
        &\le \norm{q_0 - \bar{q}^*}_\infty + \int_0^t \mre^{-(t-s)} |g^* - f(q_s)|ds\\
        &\le \norm{p_0 - \bar{q}^*}_\infty + \int_0^t \mre^{-(t-s)} |f(\bar{q}^*) - f(q_s)| ds\\
        &\le (1+\tilde{L}(1-\mre^{-t}))\norm{p_0 - \bar{q}^*}_\infty.
    \end{align*}
    Hence for any fixed $t > 0$ and any $\epsilon > 0$, we can always make $\norm{p_0 - \bar{q}^*}_\infty < \delta$ where $\delta < \frac{\epsilon}{1+\tilde{L}(1-\mre^{-t})}$ to guarantee that $\norm{p_t - \bar{q}^*}_\infty < \epsilon$. The Lyapunov stability holds, completing the proof.
\end{proof}

\subsubsection{Condition \ref{cond:ii:Hinf}}

We now look at condition \ref{cond:ii:Hinf}. 
\begin{lemma}\label{lemma-RinfinityDoeblin}
    Under assumptions \ref{assump-AROE-Doeblin} and \ref{assump-asymptcoherent} on $\mcR$, the risk map $\mcR^\infty$ also satisfies Assumption \ref{assump-AROE-Doeblin}.
\end{lemma}
\begin{proof}
    Following Assumption \ref{assump-AROE-Doeblin}, there exists a coherent risk measure $\nu$ and $\bar{\alpha} \in (0,1)$ such that for any $v \ge v' \in \mcL(\mcX)$, we have
    \begin{align*}
        \min_{(x,a)\in\mcK}\{ \mcR(v|x,a) - \bar{\alpha} \nu(v) - \mcR(v'|x,a) + \bar{\alpha} \nu(v')\} \ge 0. 
    \end{align*}
    Substituting $v$ and $v'$ with $sv$ and $sv'$ respectively, where $s > 0$, and then dividing both sides by $s$, we obtain
    \begin{align*}
        \frac{1}{s}\min_{(x,a)\in\mcK}\{ \mcR(sv|x,a) - \bar{\alpha} \nu(sv) - \mcR(sv'|x,a) + \bar{\alpha} \nu(sv')\} \ge 0.
    \end{align*}
    Since $\nu$ is coherent and by Assumption \ref{assump-asymptcoherent}, $\lim_{s\to\infty}\frac{1}{s}\mcR_{x,a}(sv) = \mcR_{s,a}^\infty(v)$, taking the limit, we obtain
    \begin{align*}
        0&\le \lim_{s\to\infty} \min_{(x,a)\in\mcK} \left\{ \frac{1}{s} \mcR(sv|x,a) - \bar{\alpha}\nu(v) - \frac{1}{s} \mcR(sv'|x,a) + \bar{\alpha}\nu(v') \right\}\\
        &=\min_{(x,a)\in\mcK} \{ \mcR_{x,a}^\infty(v) - \bar{\alpha} \nu(v) - \mcR_{x,a}^\infty(v') + \bar{\alpha} \nu(v') \}.
    \end{align*}
    This implies that $\mcR^\infty$ satisfies Assumption \ref{assump-AROE-Doeblin} with coherent risk measure $\nu$ and $\bar{\alpha} \in (0,1)$.
\end{proof}

\begin{lemma}\label{lemma-ODE-originGASE}
    Under assumptions \ref{assump-AROE-Doeblin}, \ref{assump-RQI-f}, \ref{assump-asymptcoherent} and \ref{assump-fhomogeneous}, the limit $H_\infty(q) := \lim_{s\to\infty} H_s(q)$ exists for all $Q\in\mcL(\mcK)$, and convergence is uniform on any compact sets. Furthermore, the ODE \eqref{eq-ODE-phi_infinity} has the origin as a unique globally asymptotically stable equilibrium.
\end{lemma}
\begin{proof}
    Under Assumption \ref{assump-fhomogeneous}, we have
    \begin{align*}
        H_s(q)(x,a) &= \frac{1}{s} \left\{\mcR_{x,a}\left(c(x,a) + \min_{a'\in\mcA} sq(\cdot,a')\right) - f(sq) - sq(x,a)\right\}\\
        &= \frac{c(x,a)}{s} + \frac{1}{s}\mcR_{x,a}\left(s\min_{a'\in\mcA} q(\cdot,a')\right) - f(q) - q(x,a).
    \end{align*}
    Hence, by Assumption \ref{assump-asymptcoherent}:
    \[ H_\infty(q)(x,a) := \lim_{s\to\infty} H_s(q)(x,a) = \mcR_{x,a}^\infty\left(\min_{a'\in\mcA} q(\cdot,a')\right) - f(q) - q(x,a). \]
    Since $\mcR_{x,a}^\infty$ exists and the convergence is uniform on all compact subsets of $\mcL(\mcX)$, the first part follows.
    Clearly, the origin is an equilibrium point of the ODE \eqref{eq-ODE-phi_infinity}. Following Lemma \ref{lemma-RinfinityDoeblin}, $\mcR_{x,a}^\infty$ satisfies Assumption \ref{assump-AROE-Doeblin}, then by Lemma \ref{lemma-ODE-GASE}, the origin is also the globally asymptotically stable equilibrium for ODE \eqref{eq-ODE-phi_infinity}. Hence condition \ref{cond:ii:Hinf} holds.
\end{proof}

\subsubsection{Convergence of RVI Q-learning}
We are now ready to prove Theorem \ref{thm-RVIQ-generalR}.
\begin{proof}[Proof of Theorem \ref{thm-RVIQ-generalR}]
    The almost sure boundedness condition  \ref{cond:v:bounded} is derived from Theorem 2.1 of \cite{Borkar2000sicon}, which necessitates verifying conditions  \ref{cond:i:Lipschitz}, \ref{cond:ii:Hinf}, \ref{cond:iii:martingale} and \ref{cond:iv:stepsize}. These conditions are confirmed using Lemmas \ref{lemma-HLipschitz}, \ref{lemma-ODE-originGASE}, \ref{lemma-martingale}, and Assumption \ref{assump-RobbinsMonro}. The convergence and optimality of the stochastic approximation then follow from Theorem 2.2 of \cite{Borkar2000sicon}, where conditions \ref{cond:i:Lipschitz}, \ref{cond:iii:martingale}, \ref{cond:iv:stepsize}, and \ref{cond:vi:stable} are validated through Lemmas \ref{lemma-HLipschitz}, \ref{lemma-martingale}, \ref{lemma-ODE-GASE}, and Assumption \ref{assump-RobbinsMonro}.

    Regarding the almost sure convergence of $\pi_n\to \pi^*$, one can first observe that $Q^*:\mcK\to \mcD\subset \mbR$, for some discrete set $\mcD$ with $|\mcD|\leq |\mcK|$. Letting 
    \[ \epsilon=\min_{(x,a),(x',a')\in\mcK:Q^*(x,a) \neq Q^*(x',a')}|Q^*(x,a)-Q^*(x',a')|>0, \]
    the almost sure convergence of $Q_n\to Q^*$ implies that there is a probability one set of trajectories $\mcQ$, with each trajectory $\{\bar{Q}_n\}\in\mcQ$ having the property that there  exists an $N \geq 0$ such that $\|\bar{Q}_n-Q^*\|_\infty\leq \epsilon/2$ for all $n\geq N$. This implies that for any $n\geq N$, 
    \[Q^*(x,a)>Q^*(x',a') \Longrightarrow \bar{Q}_n(x,a)>\bar{Q}_n(x',a') ,\quad \forall(x,a),(x',a')\in\mcK.\]
    We can therefore conclude that for all $x\in\mcX$ and for all $n\geq N$, we have
    \[ \argmin_{a\in\mcA}\bar{Q}_n(x,a)\subseteq \argmin_{a\in\mcA}Q^*(x,a),\quad \forall x\in\mcX. \]
    Thus the policy $\pi_n$ converges to some $\pi^*$ for all $\{\bar{Q}_n\}\in\mcQ$ almost surely.
\end{proof}

\subsection{Proof of Theorem \ref{thm-unbiasedriskmap}}
For notation simplicity, we write \eqref{eq-mcHQ} as 
\[ \hat{\mcH}(q) = \mcH_{\hat{P}_{N+1}^{1}}(q) + \frac{\Delta_N(q)}{p_N},\]
where
\[\Delta_N(q) := \mcH_{\hat{P}_{N+1}}(q) - \frac{1}{2}\left(\mcH_{\hat{P}_{N+1}^{E}}(q) + \mcH_{\hat{P}_{N+1}^{O}}(q)\right),\quad \forall q\in\mcL(\mcK). \]

To prove Theorem \ref{thm-unbiasedriskmap}, we invoke the concentration results under the 1-Wasserstein distance from \cite{Fournier2015ptrf}. The 1-Wasserstein distance between two probability measures $\mu$ and $\nu$ on $\mbR$ is defined as
\[ d_W(\mu,\nu) := \inf_{\psi\in\Psi(\mu,\nu)} \int |x -y| \psi(dx,dy), \]
where $\Psi(\mu,\nu)$ is the set of all joint probability distributions $\psi(x,y)$ with marginals $\mu$ and $\nu$.

\begin{lemma}[Concentration inequalities]\label{lemma-concentration}
    Given any $v\in\mcL(\mcX)$ and a $\bar{p}\in\mathscr{P}(\mcX)$, let $\hat{p}^k$ be the empirical distribution from $k$ realizations $\{x_1',x_2',\dots,x_k'\}$ drawn i.i.d. from $\bar{p}$. Then,
    \[\mbE\left[d_W\left( \sum_{x'\in\mcX} \hat{p}^k(x')\delta_{v(x')},\sum_{x'\in\mcX} \bar{p}(x')\delta_{v(x')}\right)\right]\leq \mfC_1\|v\|_\infty k^{-1/2}\]
    and
    \[    \mbE\left[d_W\left( \sum_{x'\in\mcX} \hat{p}^k(x')\delta_{v(x')},\sum_{x'\in\mcX} \bar{p}(x')\delta_{v(x')}\right)^2\right]\leq \mfC_2\|v\|_\infty^2 k^{-1}, \]
    for some constant $\mfC_1,\mfC_2>0$ independent of $v$, $\bar{p}$, $k$ and  $\delta_{v(x)}$ is the Dirac measure of $v(x)$.
\end{lemma}
\begin{proof}
    To simplify notations, we use $\hat{p}(\cdot)$ to denote $\hat{p}^k(\cdot)$.
    The first bound follows from Theorem 1 of \cite{Fournier2015ptrf}. Namely, there exists a $\bar{\mfC}_1>0$ such that: 
    \begin{align*}
        \mbE\left[d_W\left( \sum_{x'\in\mcX} \hat{p}(x')\delta_{v(x')},\sum_{x'\in\mcX} \bar{p}(x')\delta_{v(x')}\right)\right]  &\leq 2\bar{\mfC}_1\left(\sum_{x'\in\mcX} v(x')^2\bar{p}(x')\right)^{1/2}/k^{1/2}\\
        &\leq 2\bar{\mfC}_1\|v\|_\infty k^{-1/2}.
    \end{align*}

    By Lemma 5 and Proposition 10 in \cite{Fournier2015ptrf}, we have that for all $w\in\mcL(\mcX)$ with $\|w\|_\infty < 1$, there exists constants $\bar{\mfC}_2,\bar{\mfC}_3>0$ such that for all $\lambda \geq 0$: 
    \[\mbP\left(d_W\left( \sum_{x'\in\mcX} \hat{p}(x')\delta_{w(x')},\sum_{x'\in\mcX} \bar{p}(x')\delta_{w(x')}\right) \geq \lambda \right) \leq \bar{\mfC}_2\exp(-\bar{\mfC}_3 k \lambda^2),\]
    given the fact that $\bar{p}$ is a distribution on a finite set.
    We can thus derive that:
    \begin{align*}
        &\mbE\left[d_W\left( \sum_{x'\in\mcX} \hat{p}(x')\delta_{v(x')},\sum_{x'\in\mcX} \bar{p}(x')\delta_{v(x')}\right)^2\right] \\
        &\leq \mbE\left[4\|v\|_\infty^2 d_W\left( \sum_{x'\in\mcX} \hat{p}(x')\delta_{w(x')},\sum_{x'\in\mcX} \bar{p}(x')\delta_{w(x')}\right)^2\right]\\
        &=4\|v\|_\infty^2 \int_0^\infty \mbP\left(d_W\left( \sum_{x'\in\mcX} \hat{p}(x')\delta_{w(x')},\sum_{x'\in\mcX} \bar{p}(x')\delta_{w(x')}\right)^2\geq \lambda \right) d \lambda\\
        &\leq 4\|v\|_\infty^2 \int_0^\infty \bar{\mfC}_2\exp(-\bar{\mfC}_3 k \lambda) d \lambda = \frac{4\|v\|_\infty^2\bar{\mfC}_2}{\bar{\mfC}_3 k},
    \end{align*}
    where $w := (1/2)v/\|v\|_\infty$ is such that $\|w\|_\infty\leq 1/2 <1$.
\end{proof}

\begin{proof}[Proof of Theorem \ref{thm-unbiasedriskmap}]
    Assumption \ref{assump-WassLipschitz} implies that $\mcR_{x,a}$ is law invariant, so that $\mcR_{x,a}(v) = \varrho(\sum_{x'\in\mcX} P(x'|x,a) \delta_{v(x')})$, with $\varrho_{x,a}$ as the distribution-based risk measure associated to $\mcR_{x,a}$. Let $\hat{\mcR}_{x,a}^k$ capture the empirical risk map that employs the same distribution-based risk measure $\varrho_{x,a}$ of $\mcR_{x,a}$, but on $\sum_{x'\in\mcX}\hat{P}_k(x'|x,a)\delta_{v(x')}$ instead of $\sum_{x'\in\mcX}P(x'|x,a)\delta_{v(x')}$, where $\hat{P}_k$ stands for the empirical distribution using $2^k$ number of samples. We start by establishing two important properties of how $\mcH_{\hat{P}_{k}}(q)$ differs from $\mcH(q)$. 

    The first property consists of a bound on the expected absolute difference between $\mcH_{\hat{P}_{k}}(q)$ and $\mcH(q)$, where the expectation is taken with respect to the sampling process. Namely, for all $q \in \mcL(\mcK)$,
    \begin{align*}
    &\mbE\left[|\mcH_{\hat{P}_{k}}(q)(x,a)  - \mcH(q)(x,a)|\right]\\
    &= \mbE\left[\left|\hat{\mcR}_{x,a}^k\left(c(x,a)+\max_{a'\in\mcA}q(\cdot,a')\right)-\mcR_{x,a}\left(c(x,a)+\max_{a'\in\mcA}q(\cdot,a')\right) \right|\right]\\
    &\leq \mbE\left[\mathfrak{L}d_W\left( \sum_{x'\in\mcX} \hat{P}_k(x'|x,a)\delta_{v(x')},\sum_{x'\in\mcX} P(x'|x,a)\delta_{v(x')}\right) \right]\\
    &\leq \mathfrak{L}\mfC_1\|c(x,a)+\max_{a'\in\mcA}q(\cdot,a')\|_\infty 2^{-k/2}
    \leq \mathfrak{L}\mfC_1(2\bar{C}+\norm{q}_\infty) 2^{-k/2},
\end{align*}
where $v(x'):=c(x,a)+\max_{a'\in\mcA}Q(x',a')$. The first inequality follows from Assumption \ref{assump-WassLipschitz}, and the second one is from Lemma \ref{lemma-concentration}.

Following a similar procedure, we have the second one, which bounds the expected square difference:
\begin{align*}
    &\mbE\left[\left(\mcH_{\hat{P}_{k}}(q)(x,a)  - \mcH(q)(x,a)\right)^2 \right]\\
    &= \mbE\left[\left(\hat{\mcR}_{x,a}^k(c(x,a)+\max_{a'\in\mcA}q(\cdot,a'))-\mcR_{x,a}(c(x,a)+\max_{a'\in\mcA}q(\cdot,a'))\right)^2 \right]\\
    &\leq \mbE\left[\mathfrak{L}^2 d_W\left( \sum_{x'\in\mcX} \hat{P}_k(x'|x,a)\delta_{v(x')},\sum_{x'\in\mcX} P(x'|x,a)\delta_{v(x')}\right)^2 \right]\\
    &\leq \mathfrak{L}^2\mfC_2\|c(x,a)+\max_{a'\in\mcA}q(\cdot,a')\|_\infty 2^{-k}
    \leq \mathfrak{L}^2\mfC_2(2\bar{C}+\norm{q}_\infty) 2^{-k},
\end{align*}
where the second inequality is from Lemma \ref{lemma-concentration}.

We are now ready to show that $\mbE[\hat{\mcH}(q)] = \mcH(q)$, for all $q\in\mcL(\mcK)$, which goes as
    \begin{align*}
        \mbE&[\hat{\mcH}(q)] = \mbE\left[\mcH_{\hat{P}_{N+1}^{1}}(q) + \frac{\Delta_N(q)}{p_N}\right]\\
        &= \mbE[\mcH_{\hat{P}_{N+1}^{1}}(q)] + \sum_{k=0}^\infty \mbP(N=k)\mbE\left[\frac{\Delta_k(q)}{p_k}\big|N=k\right]\\
        &= \mbE[\mcH_{\hat{P}_{1}^{1}}(q)] + \sum_{k=0}^\infty \mbE[\Delta_k(q)]\\
        &= \mbE[\mcH_{\hat{P}_{1}^{1}}(q)] + \sum_{k=0}^\infty \mbE\left[\mcH_{\hat{P}_{k+1}}(q) 
        - \frac{1}{2}\left(\mcH_{\hat{P}_{k+1}^{E}}(q) + \mcH_{\hat{P}_{k+1}^{O}}(q)\right) \right]\\   
        &= \mbE[\mcH_{\hat{P}_{1}^{1}}(q)] + \sum_{k=0}^\infty \left( \mbE[\mcH_{\hat{P}_{k+1}}(q)]  
        - \frac{1}{2}\left(\mbE[\mcH_{\hat{P}_{k+1}^{E}}(q)] + \mbE[\mcH_{\hat{P}_{k+1}^{O}}(q)]\right)\right)\\       
        &= \mbE[\mcH_{\hat{P}_{1}^{1}}(q)] + \sum_{k=0}^\infty \left( \mbE[\mcH_{\hat{P}_{k+1}}(q)]  
        - \mbE[\mcH_{\hat{P}_{k}^{E}}(q)]\right)\\
        &=\lim_{k\rightarrow \infty}\mbE[\mcH_{\hat{P}_{k}}(q)]  = \mcH(q),
    \end{align*}
where the limit is known to exist and identified as $\mcH(q)$ since for all $(x,a)\in\mcK$ we have
\begin{align*}
    \left|\mbE[\mcH_{\hat{P}_{k}}(q)(x,a)]  - \mcH(q)(x,a)\right| & \leq \mbE[|\mcH_{\hat{P}_{k}}(q)(x,a) - \mcH(q)(x,a)|]\\
    &\leq \mathfrak{L}\mfC_1(2\bar{C}+\|q\|_\infty) 2^{-k/2},
\end{align*}
thus implying that $\norm{\mbE[\mcH_{\hat{P}_{k}}(q)]  - \mcH(q)}_\infty \to 0$ as $k \to \infty$.

We turn to bounding $\Var[\hat{\mcH}(q)(x,a)]$. Since for all $(x,a)\in\mcK$, we have
\begin{align*}
    \Var[\hat{\mcH}(q)(x,a)] = \mbE[(\hat{\mcH}(q)(x,a))^2] - (\mbE[\hat{\mcH}(q)(x,a)])^2 
    = \mbE[(\hat{\mcH}(q)(x,a))^2] - (\mcH(q)(x,a))^2,
\end{align*}
and it is known that 
$\norm{(\mcH(q))^2}_\infty \le (2\bar{C} + \norm{q}_\infty)^2 \le 8\bar{C}^2 + 2\norm{q}_\infty^2$. The remaining question is to bound $\mbE[(\hat{\mcH}(q)(x,a))^2]$.

We first give a bound on $\mbE[(\Delta_k(q)(x,a))^2]$. Notice that
\begin{align*}
    &\mbE[\Delta_k(q)(x,a)^2]
    =\mbE\left[\left(\mcH_{\hat{P}_{k+1}}(q)(x,a) 
    - \frac{1}{2}\left(\mcH_{\hat{P}_{k+1}^{E}}(q)(x,a) + \mcH_{\hat{P}_{k+1}^{O}}(q)(x,a)\right)\right)^2\right]\\
    &=\mbE\left[\left((\mcH_{\hat{P}_{k+1}}(q)(x,a)-\mcH(q)(x,a)) 
    - \frac{1}{2}\left(\mcH_{\hat{P}_{k+1}^{E}}(q)(x,a) + \mcH_{\hat{P}_{k+1}^{O}}(q)(x,a) - 2\mcH(q)(x,a)\right)\right)^2\right]\\
    &\leq \mbE[2(\mcH_{\hat{P}_{k+1}}(q)(x,a)-\mcH(q)(x,a))^2 +(\mcH_{\hat{P}_{k+1}^{E}}(q)(x,a)-\mcH(Q_n)(x,a))^2 \\
    &\qquad + (\mcH_{\hat{P}_{k+1}^{O}}(q)(x,a)-\mcH(q)(x,a))^2]\\
    &= 2\mbE[(\mcH_{\hat{P}_{k+1}}(q)(x,a)-\mcH(q)(x,a))^2] +2\mbE[(\mcH_{\hat{P}_{k}}(q)(x,a)-\mcH(q)(x,a))^2]\\
    &\leq 2\mathfrak{L}^2\mfC_2(2\bar{C}+\|q\|_\infty)^2 2^{-k-1} + 2\mathfrak{L}^2\mfC_2(2\bar{C}+\|q\|_\infty)^2 2^{-k}\\
    &= 3\mathfrak{L}^2\mfC_2(2\bar{C}+\|q\|_\infty)^2 2^{-k}.
\end{align*}

Now we are ready to derive the bound for $\mbE[(\hat{\mcH}(q)(x,a))^2]$, for any fixed $(x,a) \in \mcK$. Namely, from definition, we have
\begin{align*}
    \mbE[(\hat{\mcH}(q)(x,a))^2] 
    &\leq 2\mbE[(\mcH_{\hat{P}_{N+1}^{1}}(q)(x,a))^2] + 2\mbE\left[\left(\frac{\Delta_N(q)(x,a)}{p_N}\right)^2\right]\\
    &= 2(2\bar{C}+\|q\|_\infty)^2 + 2\mbE\left[\sum_{k=0}^\infty p_k \mbE\left[\left(\frac{\Delta_N(q)(x,a)}{p_N}\right)^2 \Big| N=k\right]\right]\\
    &= 2(2\bar{C}+\|Q_n\|_\infty)^2 + 2\sum_{k=0}^\infty \frac{1}{p_k} \mbE[\Delta_k(q)(x,a)^2]\\
    &\leq 2(2\bar{C}+\|q\|_\infty)^2 + 2\sum_{k=0}^\infty \frac{1}{p_k} 3\mathfrak{L}^2\mfC_2(2\bar{C}+\|q\|_\infty)^2 2^{-k}\\
    &= 2(2\bar{C}+\|q\|_\infty)^2 + 6\mathfrak{L}^2\mfC_2(2\bar{C}+\|q\|_\infty)^2r^{-1}\sum_{k=0}^\infty (2(1-r))^{-k}\\
    &= 2(2\bar{C}+\|q\|_\infty)^2 + 6\mathfrak{L}^2\mfC_2(2\bar{C}+\|q\|_\infty)^2 r^{-1}(1-(2(1-r))^{-1})^{-1}\\
    &= (2+6\mathfrak{L}^2\mfC_2 r^{-1}(1-(2(1-r))^{-1})^{-1})(2\bar{C}+\|q\|_\infty)^2 \\    
    &\leq (2+6\mathfrak{L}^2\mfC_2 r^{-1}(1-(2(1-r))^{-1})^{-1})(8\bar{C}^2+2\|q\|_\infty^2),   
\end{align*}
where to ensure $\sum_{k=0}^\infty 2(1-r)^{-k}$ is finite, we require $r\in(0,1/2)$. 
This implies that there exists a uniform bound $C > 0$ such that $\Var[\hat{\mcH}(q)] = \mbE[(\hat{\mcH}(q)(x,a))^2] - (\mcH(q)(x,a))^2 \le C(1+\norm{q}_\infty^2)$.
This completes the proof.
\end{proof}

\subsection{Proof of Theorem \ref{thm-UBSROCEproperties}}\label{app:thmUBSROCESRM}

We impose the following general assumption on the convexity of the loss function $\ell$.
\begin{assumption}[Convexity]\label{assump-convexity}
    The loss function $\ell(x)$ is either convex or concave on $x \ge 0$ and either convex or concave on $x\le 0$.
\end{assumption}

We prove Theorem \ref{thm-UBSROCEproperties} by proving lemmas \ref{lemma-UBSR-property}, \ref{lemma-OCE-property} and \ref{lemma-Spectral-property}.

\begin{lemma}\label{lemma-UBSR-property}
    A risk map $\mcR$ that employs a UBSR measure satisfies assumptions \ref{assump-AROE-Doeblin}, \ref{assump-asymptcoherent} and \ref{assump-WassLipschitz}, if the Markov chain satisfies Assumption \ref{assump-ergodicity} and the loss function satisfies assumptions \ref{assump-boundedslope} and \ref{assump-convexity}.
\end{lemma}

The following property of UBSR is useful as it establishes a connection between the UBSR measure and the expected utility.
\begin{lemma}[Proposition 4.113, \cite{Follmer2016}]\label{lemma-Follmer}
    Given some random variable $v$ and some $m^*\in\mbR$, the following statements are equivalent: (i) $\SR(v) = m^*$; (ii) $\mbE[\ell(v-m^*)] = 0$.
\end{lemma}

\begin{lemma}\label{lemma-UBSR-doeblin}
    Under assumptions \ref{assump-ergodicity}, \ref{assump-boundedslope}, the UBSR satisfies Assumption \ref{assump-AROE-Doeblin}.
\end{lemma}
\begin{proof}
    By Assumption \ref{assump-ergodicity}, there exists a state $\bar{x} \in \mcX$ such that $P(\bar{x}|x,a) > 0$ for all $(x,a) \in \mcK$. Let $\nu(v) := v(\bar{x})$, which trivially satisfies $\nu(\bm{0})=0$ and is coherent. Meanwhile, choose $0<\bar{\alpha}< \frac{\epsilon_1}{L_1} \min_{(x,a)\in\mcK}P(\bar{x}|x,a)\in(0,\,1)$.
    
    Given any $v \ge v' \in \mcL(\mcX)$, Lemma \ref{lemma-Follmer} implies that $\sum_{y\in\mcX} P(y|x,a) \ell(v(y) - \SR_{x,a}(v)) = 0$, and similarly for $v'$.
    We therefore have 
    \begin{align*}
        0 &= \sum_{y\in\mcX}P(y|x,a) \left(\ell(v(y) - \SR_{x,a}(v)) - \ell(v'(y) - \SR_{x,a}(v'))\right)\\
        &=\sum_{y\in\mcX}P(y|x,a) \delta(v,v',x,a,y) ((v(y) - \SR_{x,a}(v)) - (v'(y) - \SR_{x,a}(v'))),
    \end{align*}
    for some $\delta(v,v',x,a,y) \in [\epsilon_1,L_1]$ due to  Assumption \ref{assump-boundedslope}. Hence,
    \begin{equation*}
    \begin{aligned}
         &(\SR_{x,a}(v) - \SR_{x,a}(v')) \sum_{y\in\mcX}P(y|x,a) \delta(v,v',x,a,y)\\
        &\qquad = \sum_{y\in\mcX}P(y|x,a) \delta(v,v',x,a,y) (v(y) - v'(y)).
    \end{aligned}
    \end{equation*}
    Therefore, we have
    \begin{equation*}
        \begin{aligned}
            \SR_{x,a}(v) - \SR_{x,a}(v') &\ge \inf_{\begin{subarray}{c}
            \tilde{\delta}\in\mcL(\mcK\times\mcX):\tilde{\delta}(x,a,y) \in [\epsilon_1,L_1],\\
            (x,a,y) \in \mcK\times \mcX
            \end{subarray}} \frac{\sum_{y\in\mcX}P(y|x,a) \tilde{\delta}(x,a,y) (v(y) - v'(y))}{\sum_{y\in\mcX}P(y|x,a) \tilde{\delta}(x,a,y)} \\
            & \ge \frac{\epsilon_1}{L_1} \sum_{y\in\mcX}P(y|x,a) (v(y) - v'(y)),
        \end{aligned}
    \end{equation*} 
    given that $v\geq v'$.
    Hence, we have for all $(x,a)\in\mcK$,
    \begin{align*}
        &\SR_{x,a}(v) - \bar{\alpha}\nu(v) - \SR_{x,a}(v') + \bar{\alpha}\nu(v') \\
        &\geq \left(\frac{\epsilon_1}{L_1} \sum_{y\in\mcX}P(y|x,a)(v(y) - v'(y))\right) - \bar{\alpha}\left(\nu(v)-\nu(v')\right)\\
        &= \left(\frac{\epsilon_1}{L_1} \sum_{y\in\mcX}P(y|x,a)(v(y) - v'(y))\right) - \bar{\alpha}\left(v(\bar{x})-v'(\bar{x})\right)\\
        &\ge \left(\frac{\epsilon_1}{L_1} \min_{(x,a)\in\mcK}P(\bar{x}|x,a)(v(\bar{x}) - v'(\bar{x}))\right) - \bar{\alpha}\left(v(\bar{x})-v'(\bar{x})\right)\\        
        &=\left(\frac{\epsilon_1}{L_1} \min_{(x,a)\in\mcK}P(\bar{x}|x,a)-\bar{\alpha}\right)(v(\bar{x})-v'(\bar{x})) \geq 0.
    \end{align*}
     This proves that Assumption \ref{assump-AROE-Doeblin} holds.
\end{proof}

\begin{lemma}\label{lemma-elluniform}
    For any loss function $\ell:\mbR \to \mbR$ with $\ell(0) = 0$ satisfying assumptions \ref{assump-boundedslope} and \ref{assump-convexity}, define $\ell_s(x):=\frac{1}{s}\ell(sx)$. We have $\ell_s(x) \to \ell_\infty(x)$ uniformly on compact sets as $s\to\infty$, for some $\ell_\infty(x)$ that satisfies assumptions \ref{assump-boundedslope} and \ref{assump-convexity}.
\end{lemma}
\begin{proof}
    By Assumption \ref{assump-boundedslope}, \(\ell(x)\) is strictly increasing. We consider the case where \(\ell(x)\) is convex. 
    In this setting, the right derivative is non-decreasing and bounded above hence the monotone convergence theorem implies that $\ell_+'(x) \to \overline{L}\leq L_1$ as $x \to \infty$, and similarly the left derivative $\ell_-'(x) \to \underline{L} \geq \epsilon$ as $x \to -\infty$, where we slightly abuse the notation and use $\ell'_+(x)$ and $\ell'_-(x)$ to denote the left and right derivative of $\ell$ at $x$. Without loss of generality, we assume that $\overline{L} = L_1$ and $\underline{L}=\epsilon_1$. 
    Define $\ell_\infty(x) := L_1 x$ for $x \ge 0$ and $\ell_\infty(x) :=\epsilon_1x$ for $x < 0$. Remember that \(\ell_s(x) := \frac{1}{s} \ell(sx)\). We start by demonstrating that $\lim_{s\to\infty} \ell_s(x) = \ell_\infty(x)$ pointwise and will follow with confirming uniform convergence on all compact sets using Ascoli-Arzelà theorem.

    Clearly, $\lim_{s\to\infty}\ell_s(0)=0 = \ell_\infty(0)$. Now consider the case of some $\bar{x} > 0$. Based on Assumption \ref{assump-boundedslope}, for all $s>0$, we must have $(1/s)\ell(s\bar{x})\leq (1/s) L_1(s\bar{x}-0) = L_1 \bar{x}$. Hence we have $\lim_{s\to\infty} \ell_s(\bar{x}) \leq L_1 \bar{x} = \ell_\infty(\bar{x})$. On the other hand, by convexity of $\ell(x)$ over $x\geq 0$, for any $\varepsilon > 0$, one can identify some $\hat{x}\geq 0$ such that $L_1 - \varepsilon/(2\bar{x})\in [\ell'_-(\hat{x}),\ell'_+(\hat{x})]$ and therefore for all $x\geq 0$, $\ell(x)\geq \ell(\hat{x})+(L_1 - \varepsilon/(2\bar{x}))(x-\hat{x})$. Thus we must have that
    \begin{align*}
        \frac{1}{s} \ell(s\bar{x}) &\ge \frac{1}{s}(\ell(\hat{x}) + (L_1-\varepsilon/(2\bar{x}))(s\bar{x} - \hat{x}))\\
        &= \frac{1}{s}(\ell(\hat{x}) + L_1s\bar{x} - L_1\hat{x} - (\varepsilon/(2\bar{x})) s\bar{x} +(\varepsilon/(2\bar{x}))\hat{x})\\
        &= L_1 \bar{x} -(\varepsilon/(2\bar{x}))\bar{x} + \frac{1}{s}\left(\ell(\hat{x}) - (L_1  - (\varepsilon/(2\bar{x})) )\hat{x}\right)\\
        &\ge L_1 \bar{x} - \varepsilon,
    \end{align*}
    as long as $s \ge 2|\ell(\hat{x}) - (L_1 - (\varepsilon/(2\bar{x}))) \hat{x}|/\varepsilon$. Hence $\lim_{s\to\infty} \ell_s(\bar{x}) \ge \ell_\infty(\bar{x})$. Combining the two results, we conclude that $\lim_{s\to\infty}\ell_s(x) = \ell_\infty(x)$ pointwise for $x\geq 0$.

    The case where $\bar{x}<0$ is treated similarly. Namely, letting $g(x):=-\ell(-x)$, we wish to show that $\lim_{s\to\infty} (1/s)g(s\bar{x}) = \epsilon_1 \bar{x}$ for all $\bar{x} > 0$, with $g(x)$ a concave function such that $\epsilon_1 \le (g(y) - g(x))/(y-x) \le L_1$ and $g_+'(x) = \ell_-'(-x) \to \epsilon_1$ as $x\to\infty$.
    We can start with a lower bound argument $(1/s)g(s\bar{x})\geq (1/s)\epsilon_1 (s\bar{x}-0)=\epsilon_1\bar{x}$. The upper bound is a consequence of the concavity of $g$, implying the existence of some $\hat{x} > 0$ such that $\epsilon_1 + (\varepsilon / (2\bar{x}))$ belongs to the interval $[g'_-(\hat{x}), g'_+(\hat{x})]$, where $g'_-(x)$ and $g'_+(x)$ denote the left and right derivatives of $g$ at $x$, respectively. This observation leads to the subsequent argument:
        \begin{align*}
        \frac{1}{s} g(s\bar{x}) &\le \frac{1}{s}(g(\hat{x}) + (\epsilon_1+\varepsilon/(2\bar{x}))(s\bar{x} - \hat{x}))\\
        &= \epsilon_1 \bar{x} +(\varepsilon/(2\bar{x}))\bar{x} + \frac{1}{s}(g(\hat{x}) - (\epsilon_1 + (\varepsilon/(2\bar{x}))) \hat{x})\\
        &\le \epsilon_1 \bar{x} + \varepsilon,
    \end{align*}
    for large enough $s$. This let us conclude that for $x < 0$ it must hold that $\lim_{s\to\infty} (1/s)\ell(sx)= \lim_{s\to\infty} -(1/s)g(-sx) = \epsilon_1 x$.
    
    Concerning the uniform convergence on compacts set, we first observe that \(\ell_s(x)\) is uniformly bounded on compact sets. Specifically, for \(x \in [x_a, x_b]\), we have    
    $|\ell_s(x)| = |\frac{1}{s} \ell(sx)| \leq \frac{1}{s} L_1(s|x|) \leq L_1\max\{|x_a|, |x_b|\} < \infty$. Moreover, by Assumption \ref{assump-boundedslope}, both \(\ell\) and \(\ell_s\) are Lipschitz, ensuring equicontinuity. Given the equicontinuity and uniform boundedness of \(\ell_s(x)\) on compact sets, along with pointwise convergence, the Ascoli-Arzelà theorem guarantees that \(\ell_s(x) \to \ell_\infty(x)\) uniformly on compact sets. To see this, the Ascoli-Arzelà theorem provides subsequential convergence \(\ell_{s_i} \to \tilde{\ell}\) uniformly for some function \(\tilde{\ell}\), where \(s_i \uparrow \infty\) is a subsequence index. Since we also have pointwise convergence \(\ell_s \to \ell_\infty\), it follows that \(\tilde{\ell} = \ell_\infty\), implying uniform convergence of \(\ell_{s_i}\) to \(\ell_\infty\).  
    Repeating this argument, we show that every subsequence \(\{\ell_{s_i}\}\) of \(\{\ell_s\}\) has a further subsequence that uniformly converges to \(\ell_\infty\). By the subsequence principle, we conclude that \(\ell_s\) converges to \(\ell_\infty\) uniformly on compact sets.

    The cases where \(\ell(x)\) is concave or combines convex (or concave) on $x\geq 0$ and concave (or convex) on \(x \leq 0\) are derived using a similar approach. Therefore, we conclude that under assumptions \ref{assump-boundedslope} and \ref{assump-convexity}, $\ell_s$ converges to $\ell_\infty$ uniformly on compact sets.
\end{proof}

\begin{lemma}\label{lemma-UBSR-asymptcoherent}
    Under assumptions \ref{assump-boundedslope}, \ref{assump-convexity}, the UBSR risk map satisfies Assumption \ref{assump-asymptcoherent}.
\end{lemma}
\begin{proof}
    We start with establishing that for all $(x,a)\in\mcK$, all $v\in\mcL(\mcX)$, and for all $s>0$, we have
    \begin{align*}
    \frac{1}{s}\SR_{x,a}^\ell(sv)&=\frac{1}{s}\inf\{t: \mbE_{x,a}[\ell(sv-t)]\leq 0\}\\
    &=\inf\{t': \mbE_{x,a}[\ell(sv-st')]\leq 0\}\\
    &=\inf\{t': (1/s)\mbE_{x,a}[\ell(s(v-t'))]\leq 0\}\\
    &=\inf\{t': \mbE_{x,a}[\ell_s(v-t')]\leq 0\}\\
    &=\SR_{x,a}^{\ell_s}(v),
    \end{align*}
    where $\ell_s(y):=(1/s)\ell(sy)$. Showing that the UBSR risk map is asymptotically coherent therefore reduces to showing that $\SR_{x,a}^{\ell_s}(v)$ converges uniformly to $\SR_{x,a}^{\ell_\infty}(v)$ on compact sets.
    
    We start with pointwise convergence of $\SR_{x,a}^{\ell_s}(v)$ to $\SR_{x,a}^{\ell_\infty}(v)$ after recalling that by Lemma \ref{lemma-Follmer},
    \[ \mbE_{x,a}[\ell_s(v-\SR_{x,a}^{\ell_s}(v))] = 0,\quad \mbE_{x,a}[\ell_\infty(v - \SR_{x,a}^{\ell_\infty}(v))] = 0, \quad v\in \mcL(\mcX), (x,a) \in \mcK. \]
    Specifically, given any $\bar{v}\in\mcL(\mcX)$, we can define the compact set $\mcV:=[-2\|\bar{v}\|_\infty,2\|\bar{v}\|_\infty]$.
    The uniform convergence of \(\ell_s\) to \(\ell_\infty\) on compact sets (see Lemma \ref{lemma-elluniform}) implies that for any arbitrarily small \(\varepsilon > 0\), there exists a sufficiently large \(\bar{s}\) such that we have $|\ell_\infty(y) - \ell_{\bar{s}}(y)| \leq \varepsilon\epsilon_1$, for all $y\in\mcV$. Given that $|\SR_{x,a}^{\ell_\infty}(\bar{v})|\leq \|\bar{v}\|_\infty$ for all $(x,a)\in\mcK$, denoting $\tilde{\varepsilon}(y):=\ell_\infty(y) - \ell_{\bar{s}}(y)$, we have
    \begin{align*}
        0 &= \mbE_{x,a}[\ell_\infty(\bar{v} - \SR_{x,a}^{\ell_\infty}(\bar{v}))] = \sum_{x'\in\mcX}P(x'|x,a)[\ell_{\bar{s}}(\bar{v}(x') - \SR_{x,a}^{\ell_\infty}(\bar{v})) + \tilde{\varepsilon}(\bar{v}(x')-\SR_{x,a}^{\ell_\infty}(\bar{v}))]\\
        &\le \sum_{x'\in\mcX}P(x'|x,a)[\ell_{\bar{s}}(\bar{v}(x') - \SR_{x,a}^{\ell_\infty}(\bar{v})) + \varepsilon\epsilon_1] \\
        &\le \sum_{x'\in\mcX}P(x'|x,a)[\ell_{\bar{s}}(\bar{v}(x') - \SR_{x,a}^{\ell_\infty}(\bar{v}) + \varepsilon)]\\
        & = \mbE_{x,a}[\ell_{\bar{s}}(\bar{v} -\SR_{x,a}^{\ell_\infty}(\bar{v}) + \varepsilon))],
    \end{align*}
    where the last inequality comes from the fact that $\ell_{\bar{s}}(x + \varepsilon) - \ell_{\bar{s}}(x) = \frac{1}{s} \left( \ell(sx + s\varepsilon) - \ell(sx) \right) \ge \frac{1}{s} \epsilon_1 s\varepsilon = \epsilon_1\varepsilon$ due to Assumption \ref{assump-boundedslope}.
    Similarly, we have
    \[ 0 = \mbE_{x,a}[\ell_\infty(\bar{v} - \SR_{x,a}^{\ell_\infty}(\bar{v}))] \ge \mbE_{x,a}[\ell_{\bar{s}}(\bar{v}-\SR_{x,a}^{\ell_\infty}(\bar{v}) - \varepsilon)]. \]
    This implies that
    \[   \mbE_{x,a}[\ell_{\bar{s}}(\bar{v}-\SR_{x,a}^{\ell_\infty}(\bar{v}) - \varepsilon)] \le 0 \le \mbE_{x,a}[\ell_{\bar{s}}(\bar{v} -\SR_{x,a}^{\ell_\infty}(\bar{v}) + \varepsilon))]. \]
    By the monotonicity of $\ell_{\bar{s}}$ and Lemma \ref{lemma-Follmer}, we conclude that
    \[ \SR_{x,a}^{\ell_{\bar{s}}}(\bar{v}) \in [\SR_{x,a}^{\ell_\infty}(\bar{v}) - \varepsilon, \SR_{x,a}^{\ell_\infty}(\bar{v}) + \varepsilon]. \]
    This implies that $\SR_{x,a}^{\ell_{s}}(v) \to \SR_{x,a}^{\ell_\infty}(v)$ as $s\to\infty$. Using the same argument as in Lemma \ref{lemma-elluniform}, the convergence is also uniform on compact sets. 

    Therefore, we conclude that the UBSR is asymptotically coherent, i.e. Assumption \ref{assump-asymptcoherent} holds.
\end{proof}

\begin{proof}[Proof of Lemma \ref{lemma-UBSR-property}]
    The result follows by directly applying lemmas \ref{lemma-UBSR-doeblin}, \ref{lemma-UBSR-asymptcoherent} and Lemma 15 of \cite{Prashanth2022jmlr}.
\end{proof}

\begin{lemma}\label{lemma-OCE-property}
    A risk map $\mcR$ that employs an OCE risk measure satisfies assumptions \ref{assump-AROE-Doeblin} and \ref{assump-asymptcoherent}, \ref{assump-WassLipschitz}, if the Markov chain satisfies Assumption \ref{assump-ergodicity} and the loss function satisfies assumptions \ref{assump-boundedslope} and \ref{assump-convexity}.
\end{lemma}
\begin{proof}
    To prove Assumption \ref{assump-AROE-Doeblin} notice that for any $v \ge v' \in \mcL(\mcX)$, we have
    \begin{align*}
        \OCE_{x,a}(v) - \OCE_{x,a}(v') &= \inf_{t\in\mbR}\{ \xi + \mbE_{x,a}[\ell(v-t)]\} - \inf_{t'\in\mbR}\{t' + \mbE_{x,a}[\ell(v'-t')]\}\\
        &\ge \inf_{t\in\mbR}\{ \mbE_{x,a}[\ell(v - t) - \ell(v'-t)]\}\\
        & =  \sum_{y\in\mcX}P(y|x,a) \delta(v,v',t,y)(v(y) - v'(y))\\
        &\ge \epsilon_1 \sum_{y\in\mcX}P(y|x,a) (v(y) - v'(y)),
    \end{align*}
    for some $\delta(v,v',t,y) \in [\epsilon_1,L_1]$ whose existence is guaranteed by Assumption \ref{assump-boundedslope}. We choose the coherent risk measure $\nu(v) := v(\bar{x})$ and $0<\bar{\alpha} <\epsilon_1\min_{(x,a)\in\mcK}P(\bar{x}|x,a)\in (0,1)$,  since $\epsilon_1 \le \ell_-'(0) \le 1$ from the definition of OCE. By Assumption \ref{assump-boundedslope}, for any $(x,a)\in\mcK$, we have that for all $(x,a) \in \mcK$,
    \begin{align*}
        \OCE_{x,a}(v) & - \bar{\alpha} \nu(v) - \OCE_{x,a}(v') + \bar{\alpha} \nu(v') \\
        &\ge  \epsilon_1 \sum_{y\in\mcX}P(y|x,a) (v(y) - v'(y))  - \bar{\alpha}(v(\bar{x}) - v'(\bar{x})) \\
        &\ge \left(\epsilon_1 \min_{(x,a)\in\mcK}P(\bar{x}|x,a) - \bar{\alpha}\right)(v(\bar{x}) - v'(\bar{x})) \ge 0,
    \end{align*}
     This proves Assumption \ref{assump-AROE-Doeblin}.

    To prove Assumption \ref{assump-asymptcoherent}, we start with establishing that for all $(x,a) \in \mcK$, all $v\in \mcL(\mcX)$ and for all $s > 0$ we have
    \begin{align*}
        \frac{1}{s} \OCE_{x,a}^{\ell}(sv) &= \frac{1}{s} \inf_{\xi \in \mbR} \{ \xi + \mbE_{x,a}[\ell( sv - \xi)]\}\\
        &=\inf_{\xi'\in\mbR}\{\xi' + (1/s)\mbE_{x,a}[\ell(sv - s\xi')]\}\\
        &=\inf_{\xi'\in\mbR}\{\xi' + \mbE_{x,a}[\ell_s(v - \xi')]\}\\
        &= \OCE_{x,a}^{\ell_s}(v),
    \end{align*}
    where $\ell_s(y) := (1/s)\ell(sy)$. Showing that the OCE risk map is asymptotically coherent therefore reduces to showing that $\OCE_{x,a}^{\ell_s}(v)$ converges uniformly to $\OCE_{x,a}^{\ell_\infty}(v)$ on compact sets.
    
    Notice that from Proposition 2.1 in \cite{BenTal2007mafi}, the infimum in the representation of $\OCE_{x,a}(v)$ can be attained on a member of the bounded interval supporting of the distribution $\sum_{x'}P(x'|x,a)\delta_{v(x')}$. This implies that for any fixed $\bar{v}\in\mcL(\mcX)$, there exists an optimal $\xi^*\in \mcV :=[\min_{x'\in\mcX}v(x'),\max_{x'}v(x')]$ such that $\OCE_{x,a}^{\ell}(\bar{v}) = \xi^* + \mbE_{x,a}[\ell(v-\xi^*)]$.
    Hence for any fixed $\bar{v} \in \mcL(\mcX)$, we can let $\xi_s^*$ and $\xi_\infty^* \in \mcV$ be the optimal $\xi$ for the OCE with loss function $\ell_s$ and $\ell_\infty$ such that:
    \begin{align*}
        \OCE_{x,a}^{\ell_{s}}(\bar{v}) = \xi_s^* + \mbE_{x,a}[ \ell_{s}(\bar{v} - \xi_{s}^*) ],\quad \OCE_{x,a}^{\ell_{\infty}}(\bar{v}) = \xi_\infty^* + \mbE_{x,a}[ \ell_{\infty}(\bar{v} - \xi_{\infty}^*) ],\quad (x,a) \in \mcK.
    \end{align*}
    
    Following Lemma \ref{lemma-elluniform}, we have $\ell_s(x) \to \ell_\infty(x)$ uniformly on compact sets. Hence for any arbitrarily small \(\varepsilon > 0\), 
    there exists a sufficiently large $\bar{s}$ such that $|\ell_\infty(y) - \ell_{\bar{s}}(y)| \le \varepsilon$, for all $y \in \mcV$. Denoting $\tilde{\varepsilon}(y) := \ell_\infty(y) - \ell_{\bar{s}}(y)$, we have
    \begin{align*}
        \OCE_{x,a}^{\ell_\infty}(v) &= \xi_\infty^* + \mbE_{x,a}[ \ell_{\infty}(v - \xi_{\infty}^*) ]\\
        &=\xi_\infty^* + \sum_{x'\in\mcX} P(x'|x,a) [\ell_{\bar{s}}(v(x') - \xi_{\infty}^*) + \tilde{\varepsilon}(\bar{v}(x') - \xi_\infty^*)]\\
        &\ge \xi_\infty^* + \sum_{x'\in\mcX} P(x'|x,a) [ \ell_{\bar{s}}(v(x') - \xi_{\infty}^*)- \varepsilon] \\
        &\ge \xi_{\bar{s}}^* + \mbE_{x,a}[ \ell_{\bar{s}}(v - \xi_{\bar{s}}^*)] - \varepsilon = \OCE_{x,a}^{\ell_{\bar{s}}} (v) - \varepsilon.
    \end{align*}
    Similarly,
    \begin{align*}
        \OCE_{x,a}^{\ell_{\bar{s}}}(v) &= \xi_s^* + \mbE_{x,a}[ \ell_{\bar{s}}(v - \xi_{\bar{s}}^*) ]\\
        &= \xi_{\bar{s}}^* + \sum_{x'\in\mcX} P(x'|x,a)[\ell_\infty(v(x') - \xi_{\bar{s}}^*) - \tilde{\varepsilon}(v(x') - \xi_{\bar{s}}^*)]\\
        &\ge \xi_{\bar{s}}^* + \sum_{x'\in\mcX} P(x'|x,a)[\ell_\infty(v(x') - \xi_{\bar{s}}^*)] - \varepsilon\\
        &\ge \xi_{\infty}^* + \mbE_{x,a}[\ell_\infty(v - \xi_{\infty}^*)] - \varepsilon = \OCE_{x,a}^{\ell_\infty}(v) - \varepsilon.
    \end{align*}
    Hence we have 
    \[ \OCE_{x,a}^{\ell_\infty}(v) - \varepsilon \le \OCE_{x,a}^{\ell_s}(v) \le \OCE_{x,a}^{\ell_\infty}(v) + \varepsilon,\]
    which means that $|\OCE_{x,a}^{\ell_s}(v) - \OCE_{x,a}^{\ell_\infty}(v)| < \varepsilon$. This implies $\lim_{s\to\infty} \OCE_{x,a}^{\ell_s}(v) = \OCE_{x,a}^{\ell_\infty}(v)$ pointwise. Since OCE is Lipschitz continuous and uniformly bounded on compact sets, we can apply a similar reasoning as in Lemma \ref{lemma-elluniform} to establish Assumption \ref{assump-asymptcoherent}.

    Assumption \ref{assump-WassLipschitz} follows from Lemma 12 of \cite{Prashanth2022jmlr}.
\end{proof}

\begin{lemma}\label{lemma-Spectral-property}
    A risk map $\mcR$ that employs a spectral risk measure satisfies assumptions \ref{assump-AROE-Doeblin}, \ref{assump-asymptcoherent}, \ref{assump-WassLipschitz}, if the Markov chain satisfies Assumption \ref{assump-ergodicity} and the risk spectrum $\phi(\beta) \in [\epsilon_2,L_2]$ for some $\epsilon_2 > 0$ and $L_2 < \infty$ for all $\beta \in [0,1]$.
\end{lemma}
\begin{proof}
    Since spectral risk measures are coherent, Assumption \ref{assump-asymptcoherent} holds automatically. Assumption \ref{assump-WassLipschitz} follows from Lemma 13 of \cite{Prashanth2022jmlr}. 
    We are left with Assumption \ref{assump-AROE-Doeblin}. 
    
    Notice that
    \begin{align*}
        M^\phi(v) &= \int_0^1 (\phi(\beta) - \epsilon_2) F_v^{-1}(\beta) d\beta + \epsilon_2 \int_0^1 F_v^{-1}(\beta) d\beta \\
        &= \epsilon_2 \mbE[v] + \int_0^1 \tilde{\phi}(\beta) F_v^{-1}(\beta) d\beta = \epsilon_2 \mbE[v] + M^{\tilde{\phi}}(v),
    \end{align*}
    where $\tilde{\phi}(\beta) := \phi(\beta) - \epsilon_2 \in [0,L_2-\epsilon_2]$ for all $\beta \in [0,1]$. Therefore, for any $v \ge v'\in\mcL(\mcX)$, choosing the coherent risk measure $\nu(v) = v(\bar{x})$ and setting $0<\bar{\alpha} <\epsilon_2 \min_{(x,a)\in\mcK} P(\bar{x}|x,a) \in (0,1)$, since $\epsilon_2 \le \int_0^1 \phi(\beta)d\beta = 1$, we have for any $(x,a)\in\mcK$, 
    \begin{align*}
        M^{\phi}_{x,a}(v) &- \bar{\alpha}\nu(v) - M^{\phi}_{x,a}(v') + \bar{\alpha}\nu(v') \\
        &= \epsilon_2(\mbE_{x,a}[v] - \mbE_{x,a}[v']) + M_{x,a}^{\tilde{\phi}}(v) - M_{x,a}^{\tilde{\phi}}(v') - \bar{\alpha}(\nu(v) - \nu(v'))\\
        &\ge \epsilon_2 \sum_{y\in\mcX} P(y|x,a) (v(y) - v'(y)) - \bar{\alpha}(v(\bar{x}) - v'(\bar{x}))\\
        &\ge \min_{(x,a)\in\mcK}P(\bar{x}|x,a) (\epsilon_2-\epsilon_2) (v(\bar{x}) - v'(\bar{x})) \ge 0,
    \end{align*}
    where the first inequality follows from the fact that spectral risk measures are monotone. Therefore, Assumption \ref{assump-AROE-Doeblin} holds.
\end{proof}

It is worth noting that for the widely used OCE measure CVaR, the corresponding loss function, given by \(\ell(x) = (1-\alpha)^{-1} (x)^+\), has a minimum slope of 0; its risk spectrum, defined as \(\phi(\beta) = (1-\alpha)^{-1} \bm{1}\{\beta \ge \alpha\}\), attains a minimal value of 0. 
As a result, CVaR does not satisfy the condition required in Theorem \ref{thm-UBSROCEproperties}. However, when mixed with the expectation, the mean-CVaR risk measure with $\eta > 0$ fulfills the necessary condition for spectral risk measure in Theorem \ref{thm-UBSROCEproperties} and hence satisfies Assumption \ref{assump-AROE-Doeblin}.

\begin{proof}[Proof of Theorem \ref{thm-UBSROCEproperties}]
    The result follows directly by applying lemmas \ref{lemma-UBSR-property}, \ref{lemma-OCE-property} and \ref{lemma-Spectral-property}.
\end{proof}

\newpage

\section{Additional Details and Results of Experiments}\label{sec-AdditionalExperiments}
This section presents further experiments on the convergence of the MLMC Q-learning algorithm \eqref{algo-RVIQ-generalR}, along with statistical experiments on its sample efficiency. In addition, we conduct further experiments regarding the convergence of synchronous and asynchronous UBSR Q-learning with different loss functions, along with statistical results comparing this algorithm to the MLMC Q-learning algorithm \eqref{algo-RVIQ-generalR}. Furthermore, we provide details on the application setups used in the main text. We also include a risk analysis based on the expectile parameters across different application scenarios.
All the experiments were carried out using Python 3.9 on a Linux server equipped with a 64-core AMD EPYC 7763 processor.

\subsection{Statistical Experiments on MLMC}\label{sec-StatisticalMLMC}
Although Theorem \ref{thm-unbiasedriskmap} ensures controllable variance for \( r \in (0,1/2) \), it requires an infinite number of samples in expectation per iteration to achieve this. However, our experiments indicate that controllable variance can still be attained for some $r \in (1/2,3/4)$, as demonstrated in \cite{WangSB2023aistats} for a special case of distributionally robust discounted MDP. 

Table \ref{table-MLMCStatistics} shows the statistical results (average number of samples, average estimated optimal risk and standard deviation of estimated optimal risk) from 100 simulations, each consisting of 1,000 iterations of the MLMC Q-learning algorithm \eqref{algo-RVIQ-generalR} based on MLMC, for different values of the geometric parameter $r$ changing from 0.49 to 0.9, under a randomly generated MDP with 3 states and 3 actions, following the generation procedure outlined in Section \ref{sec-Experiments-ConvergenceRVIQ}.
We observe that for small values of $r$, the number of samples required to estimate the risk measure is quite large, but it decreases as $r$ increases. Additionally, the final estimated optimal average risk closely approximates the true average risk, which is 0.2968, computed via risk-aware RVI, suggesting that the MLMC Q-learning algorithm indeed converges to the right value. The standard deviation of estimated optimal risk initially decreases starting at $r=0.49$, but begins to rise again at $r=0.70$. This supports the findings in \cite{WangSB2023aistats}, indicating that MLMC could offer finite sample guarantee and controllable variance for some $r \in (1/2, 3/4)$.

\begin{table}[htbp]
  \centering
  \caption{Statistical properties of the MLMC Q-learning algorithm for different $r$}
   \resizebox{0.75\textwidth}{!}{
    \begin{tabular}{cccc}
    \toprule
    $r$ & \textbf{Average Number} & \textbf{Average Estimated} & \textbf{Standard Deviation of} \\
     & \textbf{of Samples} & \textbf{Optimal Risk} & \textbf{Est. Opt. Risk} \\
    \midrule
    0.49  & 202615.68 & 0.2956 & 0.0159 \\
    0.50   & 174297.02 & 0.2973 & 0.0163 \\
    0.55 & 92635.72 & 0.2976 & 0.0153\\
    0.60   & 52500.64 & 0.2974 & 0.0164 \\
    0.65 & 39152.40 & 0.2963 & 0.0157\\
    0.70 & 31476.86 & 0.2983 & 0.0188\\
    0.75  & 26955.32 & 0.3002 & 0.0210 \\
    0.80   & 23967.84 & 0.2966 & 0.0221 \\
    0.90   & 20257.32 & 0.2902 & 0.0326 \\
    \bottomrule
    \end{tabular}%
    }
  \label{table-MLMCStatistics}%
\end{table}%

\subsection{Additional Experiments for UBSR Q-learning Algorithm}\label{sec-AlternativeRVIQUBSR}
For completeness, we present the synchronous version of UBSR Q-learning algorithm as follows: for all $(x,a) \in \mcK$,
\begin{equation}\label{algo-RVIQ-UBSR}
    Q_{n+1}(x,a) = Q_n(x,a) + \gamma(n) \ell \Big( c(x,a) +\min_{a' \in \mcA} Q_n(x',a') - f(Q_n) - Q_n(x,a)  \Big),
\end{equation}
where $x' \sim P(\cdot|x,a)$, $\gamma(n)$ is the step size satisfying Assumption \ref{assump-RobbinsMonro} and $f(Q_n)$ serves as the relative value satisfying Assumption \ref{assump-RQI-f} and \ref{assump-fhomogeneous}.

In addition to the expectile experiment presented in Section \ref{sec-UBSRQ-learning}, we also provide the convergence results for the special case of the synchronous UBSR Q-learning algorithms with polynomial mixed utility (also referred to as the S-shape utility, as discussed in \cite{Shen2014neuralcomp}) and the soft quantile (as discussed in \cite{Hau2024aistats}). The polynomial mixed utility function, derived from prospect theory, is defined as follows:
\begin{align*}
    \ell_{\mathrm{PM}}(x) := \begin{cases}
        k_1 x^{b_1}, & x \ge 0,\\
        -k_2 (-x)^{b_2}, & x < 0,
    \end{cases}
\end{align*}
where $k_1,k_2 > 0$, $b_1, b_2 \ge 0$. The soft quantile, used as an approximation for the quantile, is defined as follows:
\begin{align*}
    \ell_{\mathrm{SQ}}(x) := \begin{cases}
        (1-\alpha)(\kappa x + \kappa^2 - 1), & x < -\kappa,\\
        \frac{1-\alpha}{\kappa} x, & -\kappa \le x < 0,\\
        \frac{\alpha}{\kappa} x , & 0 \le x <\kappa,\\
        \alpha(\kappa x - \kappa^2 + 1), & x \ge \kappa,
    \end{cases}
\end{align*}
with $\alpha \in [0,1]$ and $\kappa > 0$.
We choose $k_1 = 1-k_2 = 0.3$, $b_1 = b_2 = 0.5$ for the polynomial mixed utility and $\alpha = 0.2$, $\kappa = 2$ for the soft quantile. It is worth noticing that both two loss functions are generally neither convex nor concave on the whole domain. 

\begin{figure}[htbp!]
    \centering
    \subfigure{
        \includegraphics[width=0.45\textwidth]{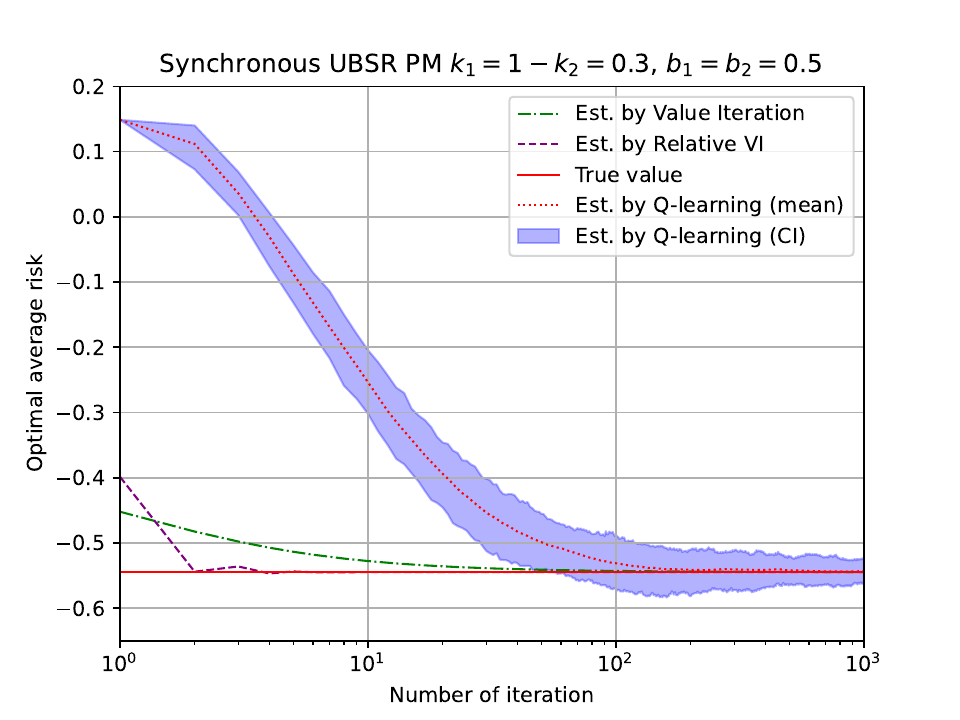}
    }
    \subfigure{
        \includegraphics[width=0.45\textwidth]{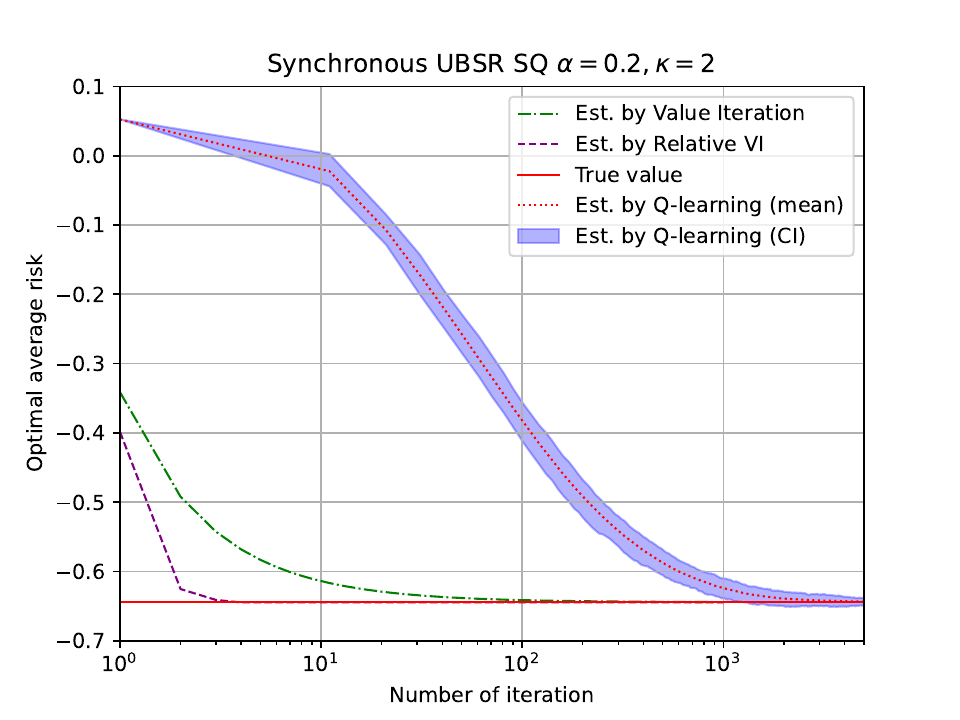}
    }
    \caption{Convergence of the synchronous UBSR Q-learning algorithm \eqref{algo-RVIQ-UBSR} for polynomial mixed utility and soft quantile.}
    \label{figure:UBSR_PM}
\end{figure}

Under the same MDP and step size settings as in Section \ref{sec-Experiments-ConvergenceRVIQ}, we run our algorithm 100 times independently and plot the trajectory obtained from the value iteration, risk-aware RVI algorithm \eqref{algo-RVI-generalR}, mean value of $f(Q)$ across all 100 trajectories from the synchronous UBSR Q-learning algorithm \eqref{algo-RVIQ-UBSR} and 95th and 5th percentiles as the upper and lower bound of the 100 trajectories as the confidence interval.
The results are presented in Figure \ref{figure:UBSR_PM}. It appears that our synchronous UBSR Q-learning algorithm successfully converges to the true optimal average risk with high probability under both instances of the loss functions.

\begin{figure}[htbp!]
    \centering
    \subfigure{
        \includegraphics[width=0.45\textwidth]{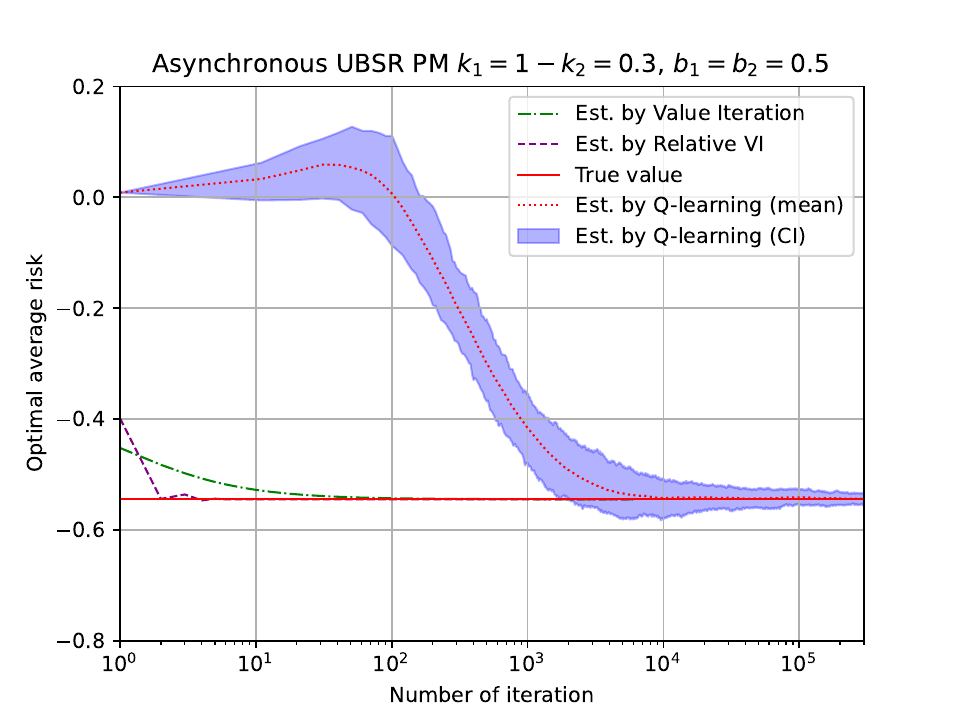}
    }
    \subfigure{
        \includegraphics[width=0.45\textwidth]{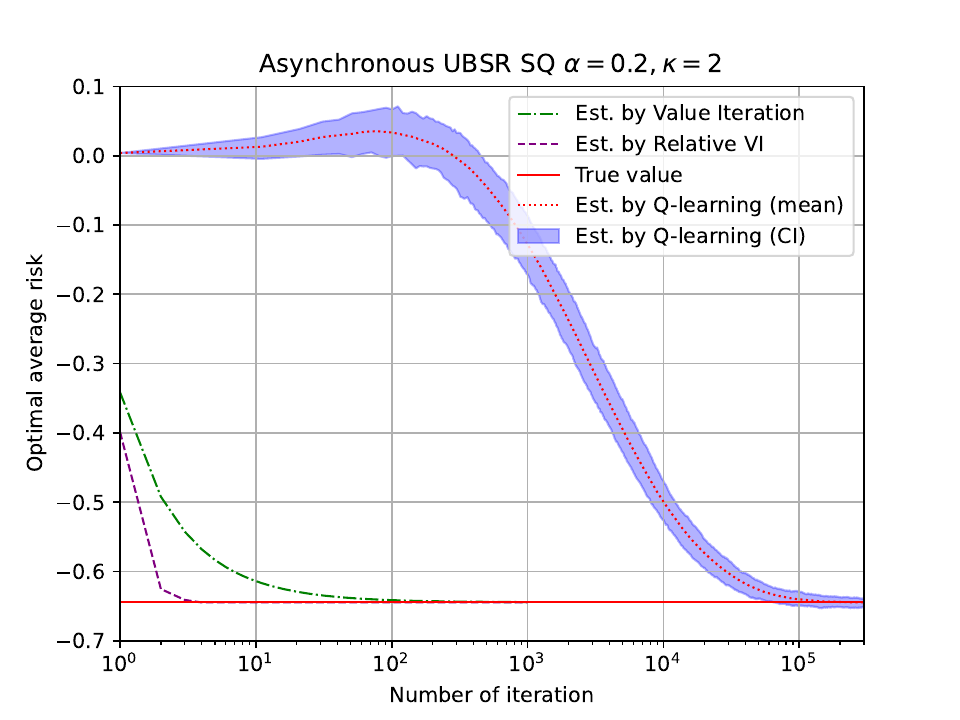}
    }
    \caption{Convergence of the asynchronous UBSR Q-learning algorithm \eqref{algo-RVIQ-UBSR-Asynchronous} for polynomial mixed utility and soft quantile.}
    \label{figure:AUBSR}
\end{figure}

Figure \ref{figure:AUBSR} presents the convergence experiments for the asynchronous UBSR Q-learning algorithm \eqref{algo-RVIQ-UBSR-Asynchronous} under the same settings as in Section \ref{sec-Experiments-ConvergenceRVIQ}, with polynomial mixed utility parameters \( k_1 = 1-k_2 = 0.3 \), \( b_1 = b_2 = 0.5 \) and soft quantile parameters $\alpha = 0.2$, $\kappa =2$, under 300,000 iterations. The results provide evidence that the asynchronous algorithm also converges with high probability, confirming the applicability of the off-policy UBSR Q-learning algorithm \eqref{algo-RVIQ-UBSR-Asynchronous}.

Table \ref{table-MLMC-UBSR} shows the statistical properties (average number of iterations, average estimated optimal risk, standard deviation of estimated optimal risk and average risk of estimated policy) of 100 simulations comparing the MLMC Q-learning algorithm \eqref{algo-RVIQ-generalR} with the synchronous UBSR (S-UBSR) Q-learning algorithm \eqref{algo-RVIQ-UBSR} and asynchronous UBSR (A-UBSR) Q-learning algorithm \eqref{algo-RVIQ-UBSR-Asynchronous} under the same settings as in Section \ref{sec-Experiments-ConvergenceRVIQ}. All the algorithms are using the equivalent number of samples. The total sample size for the MLMC-based and A-UBSR Q-learning algorithms is 300,000, equivalent to 1,000 iterations for the S-UBSR Q-learning algorithm since the S-UBSR Q-learning generates one sample per state-action pair during each iteration. For the MLMC Q-learning algorithm, we select $r > 0.5$ to ensure a finite average number of samples per iteration. The optimal average risk, computed through the risk-aware RVI \eqref{algo-RVI-generalR}, is -0.1076. Additionally, we compare the mean average risk derived from the policies produced by the algorithms to assess whether the algorithms provide the optimal policy.

\begin{table}[htbp]
  \centering
  \caption{Statistics of solutions from risk-aware RVI Q-learning algorithms after 300,000 $(x,a,x')$ observations in a setting where the true optimal average risk is -0.1076.}
  \resizebox{\textwidth}{!}{
    \begin{tabular}{cccccc}
    \toprule
    \textbf{Algorithm} & $r$    & \textbf{Average Number} & \textbf{Average Estimated } & \textbf{Standard Deviation of} & \textbf{Average Risk of} \\
     &     & \textbf{of Iterations} & \textbf{Optimal Risk} & \textbf{Est. Opt. Risk} & \textbf{Estimated Policy} \\    
    \midrule
    \multirow{7}[2]{*}{MLMC} & 0.55  & 729.16 & -0.1086 & 0.0137 & -0.1052 \\
          & 0.60   & 1039.22 & -0.1099 & 0.0103 & -0.1060 \\
          & 0.65  & 1396.86 & -0.1067 & 0.0104 & -0.1060 \\
          & 0.70   &  1731.77       & -0.1096 & 0.0099 & -0.1065 \\
          & 0.75  & 2000.43 & -0.1097 & 0.0089 & -0.1053 \\
          & 0.80   & 2250.29 & -0.1113 & 0.0114 & -0.1059 \\
          & 0.90   & 2667.43 & -0.1165 & 0.0206 & -0.1050 \\
    \midrule
    S-UBSR  & $-$ & 6000  & -0.1076 & 0.0030 & -0.1076 \\
    A-UBSR & $-$ & 300000 & -0.1074 & 0.0029 & -0.1076  \\
    \bottomrule
    \end{tabular}%
    }
  \label{table-MLMC-UBSR}%
\end{table}%

\begin{figure}[htbp!]
    \centering
    \includegraphics[width=0.5\linewidth]{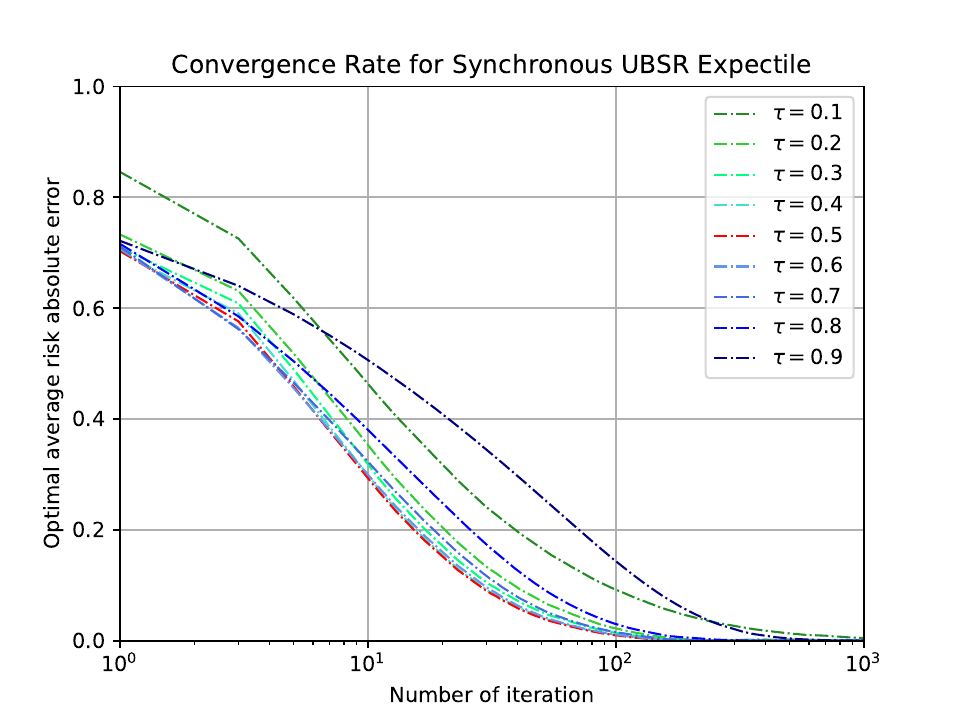}
    \caption{Convergence rate of synchronous UBSR Q-learning for expectile under different $\tau$.}
    \label{figure:UBSRconvergenceRate}
\end{figure}

Figure \ref{figure:UBSRconvergenceRate} illustrates the convergence rate of the synchronous UBSR Q-learning algorithm under the expectile risk measure for different values of $\tau$. The $y$-axis denotes the absolute error between the current value and the optimal average risk.
Recall that under the expectile risk measure, $\tau = 0.5$ represents a risk-neutral agent, $\tau > 0.5$ corresponds to a risk-averse agent, and $\tau < 0.5$ to a risk-seeking agent. The results show that convergence occurs faster for a risk-neutral agent and more slowly as the agent becomes more risk-aware. This suggests that incorporating risk-awareness in a way that is more sensitive to the tail events (i.e. $\tau$ going towards $0$ or $1$) increases the computational effort required.

From the experiments, we conclude that although constrained to the UBSR class of risk measures, the UBSR Q-learning algorithm shows significantly higher computational efficiency. It achieves notably lower standard error, faster convergence, and greater ease of implementation. 

Although the convergence experimental results are promising, analyzing the almost sure convergence of the UBSR Q-learning algorithm \eqref{algo-RVIQ-UBSR} remains a major challenge. The standard approach for proving the almost sure convergence of the average cost Q-learning algorithm relies on the ODE analysis of stochastic approximation \citep{Abounadi2001sicon, Borkar2000sicon}. In this case, the analysis leads to a high-dimensional nonlinear ODE system, which lacks the desirable property observed in the risk-neutral setting and our MLMC-based approach, where the difference between the reference ODE $q$ and the target ODE $p$, when starting from the same initial point, remains a scalar function over time. Consequently, Lemma \ref{lemma-ODE-r} does not hold.

\subsection{Application Setups}\label{sec-ApplicationSetups}
The \textit{machine replacement} problem (e.g.  Section 6.10.4, \cite{Puterman1994book}) involves managing a machine that deteriorates over time, with the goal of minimizing the long-term average cost. The machine can be in various states representing its condition, ranging from new to totally break down. At each time step, the agent chooses between two actions: keep operating the machine or replace it with a new one. Operating the machine incurs maintenance and operational costs, which increase as the machine deteriorates, while replacing it incurs a significant one-time cost but resets the machine to its best condition. 

For the parameters, we define a scenario with 30 degradation states, where state 0 represents a fully new machine and state 29 corresponds to a failure. The degradation probabilities are generated randomly, with a decreasing probability of transitioning to higher degradation states as the machine's condition worsens. Additionally, there is always a positive probability of transitioning to the failure state. The replacement cost is set to $30^{1.5}$, the operating cost is $1\times s$, and the maintenance cost is $0.5\times s^{1.5}$, where $s$ denotes the current state level. Additionally, the failure cost is twice the replacement cost, ensuring significant penalties for machine failure. 

The \textit{water reservoir management} problem (e.g.  Section 1.3, \cite{Hernandez1989book}) involves managing a reservoir to balance water supply, demand, and the risk of overflow or shortage. The reservoir has discrete states representing water levels, and at each time step, the decision-maker chooses how much water to release. The goal is to minimize the long-term average cost, which includes penalties for water shortages, overflows, and operational costs. 

For the parameters, we define the maximum water level as 19 and the maximum release as 5. The demand is set to 4, with a shortage cost of 15 per level shortage, an overflow cost of 20 per level overflow, and an operational cost of 2 per unit of water released. The probability of the incoming water level is randomly generated, with a decreasing probability of transitioning to higher water levels, reflecting the natural variability of inflows. However, there is always a positive probability of reaching the maximum water level, ensuring that the risk of overflow is accounted for in every state.

The \textit{inventory management} problem (e.g.  Section 1.3, \cite{Hernandez1989book}) involves managing stock levels to meet stochastic demand while minimizing long-term average costs. The system has discrete states representing inventory levels, and at each time step, the agent chooses how much to order to replenish stock. Costs include holding costs for inventory, ordering costs for placing orders, and shortage costs for unmet demand. 

For the parameters, we set the maximum inventory level to 9 and the maximum demand to 9. The probability of the incoming demand is generated randomly with lower probability for higher demand. The holding cost per unit of inventory is 1, the ordering cost per unit is 5, and the shortage cost per unit of unmet demand is 10.

\subsection{Risk Analysis Based on Parameter of Expectile}\label{sec-RiskAnalysisExpectile}

\begin{figure}[htbp!]
    \centering
    \subfigure{
        \includegraphics[width=0.45\textwidth]{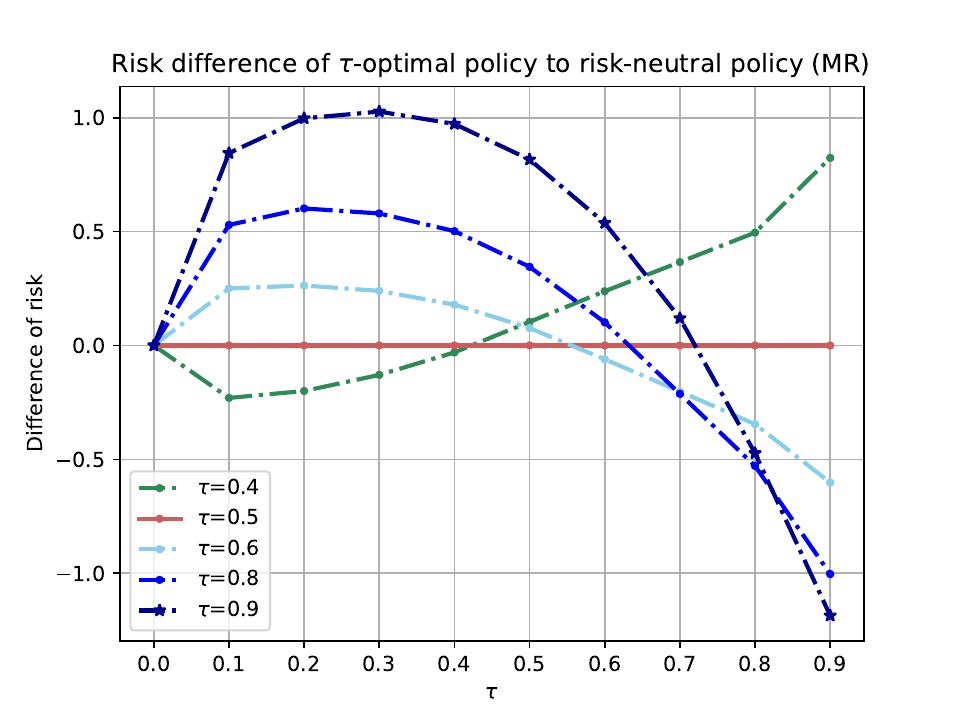}
    }
    \subfigure{
        \includegraphics[width=0.45\textwidth]{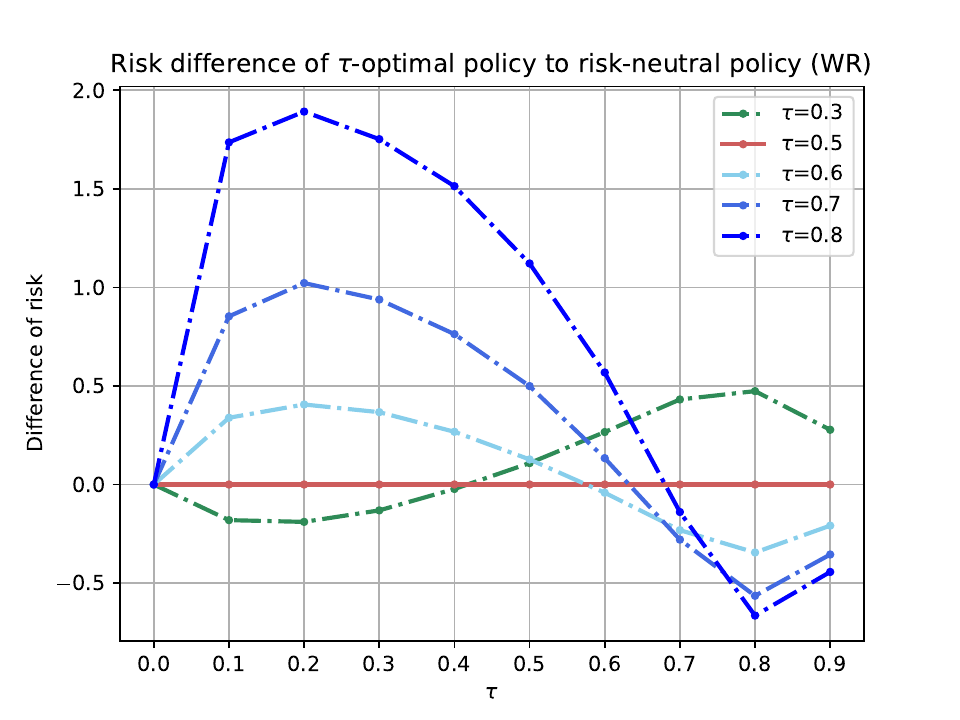}
    }
    \caption{Risk difference between $\tau$-optimal policy and risk-neutral policy under different $\tau$-values for MR and WR.}
    \label{figure:ExpectileComparison}
\end{figure}

We visualize the results of the machine replacement and water reservoir management problems under different $\tau$ parameters of the expectile in Figure \ref{figure:ExpectileComparison}. This figure illustrates the difference between the $\tau$-optimal average risk and the average risk evaluated under the risk-neutral policy. The findings confirm that for $\tau < 0.5$, the agent exhibits risk-seeking behavior, whereas for $\tau > 0.5$, the agent becomes risk-averse. Notably, when $\tau > 0.5$, the $\tau$-optimal policy achieves a lower average risk than the risk-neutral policy, reaching the minimum average risk at the corresponding $\tau$.

\begin{figure}[htbp]
    \centering
    \includegraphics[width = \textwidth]{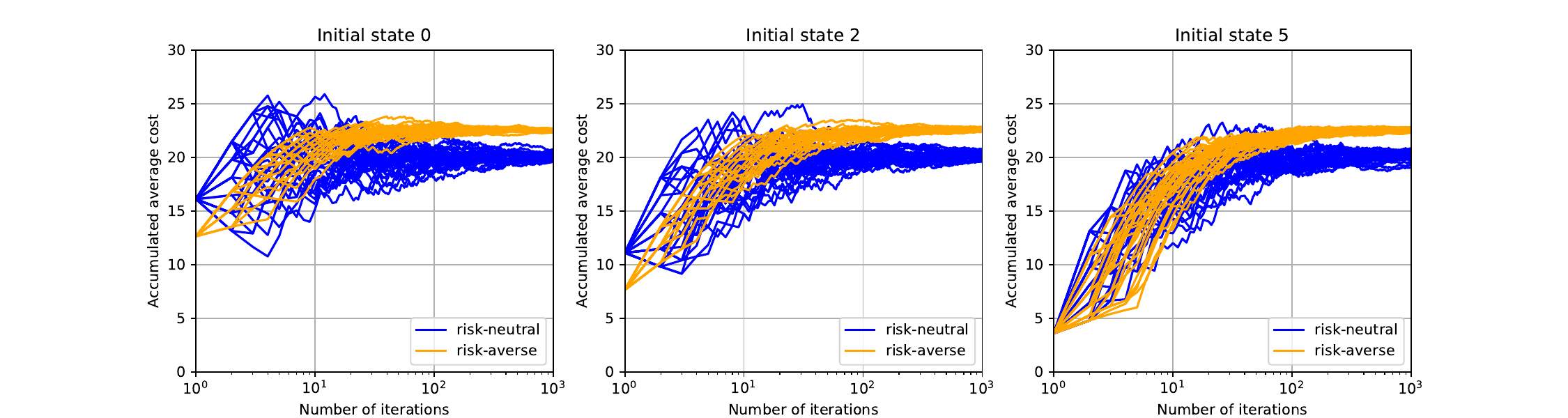}
    \caption{Simulation trajectories of risk-neutral policy and risk-averse policy under the risk-neutral setting for IM.}
    \label{figure:SimulationTrajec}
\end{figure}

In Figure \ref{figure:SimulationTrajec}, we present 30 simulation trajectories for both the risk-neutral policy and the risk-averse expectile policy (with $\tau = 0.9$) across 1,000 iterations for the inventory management problem. The risk-neutral policy results in an optimal policy of (5,4,3,2,1,0,0,0,0,0), while the risk-averse policy yields an optimal policy of (2,1,0,0,0,0,0,0,0,0), with the first element of the vector representing zero inventory. It is evident that the risk-averse policy produces trajectories with lower variance, suggesting that it could offer greater stability when observed over a shorter time frame.

\end{document}